%% file: thesis.tex
\documentclass{elteikthesis}
\usepackage{hyperref}
\hypersetup{
    colorlinks=true,
    linkcolor=blue,
    filecolor=magenta,      
    urlcolor=cyan,
}

\title{Autonomous Navigation in Dynamic Environments: Deep Learning-Based Approach} 
\date{\textbf{2020}} 

\author{\Large \textbf{Omar Mohamed \hspace{2cm} Zeyad Mohsen}\\
\hspace{1.7em} \Large \textbf{Mohamed Wageeh \hspace{1.5cm} Mohamed Hegazy}\\}

\supervisor{\Large \textbf{Dr. Mohamed S. Elbakry}} 
\affiliation{\textbf{Institute of Aviation Engineering \& Technology}} 
\extsupervisor{\Large \textbf{Prof. Moustafa Elshafei}} 
\extaffiliation{\textbf{Zewail City of Science \& Technology}} 
\cosupervisor{\Large \textbf{Mr. Ihab S. Mohamed}}
\coaffiliation{\textbf{INRIA Sophia Antipolis - Méditerranée, France}}
\university{\textbf{Institute of Aviation Engineering}\\
\textbf{and Technology}} 
\department{\textbf{Department of Electronics and} \\ \textbf{Communications Engineering}} 
\city{\large \textbf{Giza, Egypt}} 
\logo{logo.png} 

\begin{document}

\documentlang{english}


\input{settings.tex}

\maketitle

\input{chapters/acknowledgement.tex}

\input{chapters/abstract.tex}
\cleardoublepage


\tableofcontents
\cleardoublepage

\input{chapters/intro.tex}

\cleardoublepage

\input{chapters/literature-review.tex}

\cleardoublepage

\input{chapters/CNN.tex}

\cleardoublepage
\input{chapters/Preparation-Studies.tex}

\cleardoublepage
\input{chapters/Proposed_Approach_and_Simulation_Results.tex}
\cleardoublepage

\cleardoublepage
\input{chapters/Conclusions-and-Future-Work.tex}

\cleardoublepage




\addcontentsline{toc}{chapter}{\biblabel}
\printbibliography[title=\biblabel]
\cleardoublepage








\end{document}

%% file: settings.tex
\counterwithout{footnote}{chapter}

\newcounter{conpageno}
\let\oldtableofcontents\tableofcontents
\renewcommand{\tableofcontents}{
	\pagenumbering{roman}
	\oldtableofcontents
	\cleardoublepage
	\listoffigures
	\cleardoublepage
	\pagenumbering{arabic}
	\setcounter{page}{1}
}

%% file: chapters/acknowledgement.tex
\chapter*{\centering Acknowledgements}
\thispagestyle{empty} 

In the Name of ALLAH, the Most Merciful, the Most Compassionate all praise be to ALLAH and prayers and peace be upon the Prophet Mohamed. First and foremost, we are totally sure that this work would have never become truth, without help of ALLAH.

Secondly, we would like to express our deepest appreciation to our supervisors professor Mostafa Elshafei and Dr. Mohamed Sobhy for their continuous stimulating suggestions and encouragement. We would like to express our special thanks of gratitude to professor Mostafa Elshafei, who granted us the opportunity for achievement our graduation project at Zewail City. We would like to express our deepest appreciation to Dr. Mohamed Sobhy for his patient academic guidance through the research and preparation of this thesis. Because of their invaluable advices and constructive direction, I have been able to finish this dissertation for my graduation project.

Third, we would like to acknowledge with much appreciation the crucial role of Eng. Ihab S. Mohamed. This work wouldn't have been finalized at this level without his continuous support in every aspect of the project as well as his fruitful discussions and bright suggestions from the start. He has taught us so much about robotics and have helped develop a stronger desire to continue to pursue research in the field. Thanks a lot for his tediously reviewing our thesis.

Finally, we must express our very profound gratitude to our parents, sisters, brothers, and families for providing us with unfailing support and continuous encouragement throughout our years of study and through the process of working on this project. This accomplishment would not have been possible without them. Thank you. 

%% file: chapters/abstract.tex
\chapter*{\centering Abstract}
\thispagestyle{empty} 

Mobile robotics is a research area that has witnessed incredible advances for the last decades. Robot navigation is an essential task for mobile robots. Many methods are proposed for allowing robots to navigate within different environments. This thesis studies different deep learning-based approaches, highlighting the advantages and disadvantages of each scheme. In fact, these approaches are promising that some of them can navigate the robot in unknown and dynamic environments. In this thesis, one of the deep learning methods based on convolutional neural network (CNN) is realized by software implementations. There are different preparation studies to complete this thesis such as introduction to Linux, robot operating system (ROS), C++, python, and GAZEBO simulator. Within this work, we modified the drone network (namely, DroNet) approach to be used in an indoor environment by using a ground robot in different cases. Indeed, the DroNet approach suffers from the absence of goal-oriented motion. Therefore, this thesis mainly focuses on tackling this problem via mapping using simultaneous localization and mapping (SLAM) and path planning techniques using Dijkstra. Afterward, the combination between the DroNet ground robot-based, mapping, and path planning leads to a goal-oriented motion, following the shortest path while avoiding the dynamic obstacle. Finally, we propose a low-cost approach, for indoor applications such as restaurants, museums, etc, on the base of using a monocular camera instead of a laser scanner.

%% file: chapters/intro.tex
\chapter{Introduction} 
\label{ch:intro}

Recently mobile robots have started to work in the real world scenarios. Applications of mobile robots are immense and acquiring importance. These applications include agricultural robotics such as fertilizing and planting , support to medical services such as transportation of medication , client support such as museum tour, exhibition guides and military missions such as surveillance and monitoring. A group of mobile robots can do work in parallel and it has advantages over single robot systems. Multi mobile robot systems can complete a given task faster as compared to a single robot. In such tasks where multi mobile robots are involved, there is a requirement that all robots navigate and avoid each other to reach their goal positions. Multi mobile robot systems can be used for material transportation in factories, defense, agricultural robotics and service support.

\section{Motivation}\label{motivation}

Mobile robot is an autonomous agent capable of navigating intelligently anywhere using sensor-actuator control techniques. The applications of the autonomous mobile robot in many fields such as industry, space, defence and transportation, and other social sectors are growing day by day. Furthermore, navigation from one point to another point is one of the most basic tasks almost in every robotic system nowadays. There are many methods have been proposed throughout the last century to achieve this fundamental operation \cite{mohamed2020model}. Also, there are several challenges that are faced during navigation. These challenges include fluctuations in navigation accuracy depending on the complexity of the environment as well as problems in mapping precision, localization accuracy, actuators efficiency and etc. Till now, the navigation system in dynamic environments is the main important challenges in mobile robot systems. Recently, this topic is one of the hot research areas. So, there are many approaches to achieve this task with the highest possible accuracy. Thus, this thesis will focus on the deep learning approaches as they have showed the most auspicious results of all the various investigated methods.

\section{Definition of Autonomous Navigation}

Autonomous navigation means that a robot is able to plan its path and execute its
plan without human intervention. In some cases remote navigation aids are used
in the planning process, while at other times the only information available to
compute a path is based on input from sensors aboard the robot itself. An
autonomous robot is one which not only can maintain its own stability as it moves
but also can plan its movements. Autonomous robots use navigation aids when
possible but can also rely on visual, auditory, and olfactory cues. Once basic
position information is gathered in the form of triangulated signals or
environmental perception, machine intelligence must be applied to translate some basic motivation (reason for leaving the present position) into a route and motion plan. This plan may have to accommodate the estimated or communicated
intentions of other autonomous robots in order to prevent collisions, while
considering the dynamics of the robot's own movement envelope.

\section{Problem Statement}

The main core of this project is studying and evaluation the state-of-art the deep learning approaches for robotic navigation which are recently proposed in both static and dynamic environments. After studying the implementation of each approach and the advantages and disadvantages of these algorithms we were able to set our minds on studying and modifying the DroNet approach  \cite{loquercio2018dronet}. The DroNet approach was proposed to accomplish the requirements for civilian drones that are soon expected to be used in a wide variety of tasks such as aerial surveillance, delivery, or monitoring of existing architectures. The DroNet approach is a convolutional neural network (CNN) that can safely drive a drone through the streets of a city.\footnote{For supplementary video see: \url{https://youtu.be/ow7aw9H4BcA}} This approach mainly works in outdoor and indoor environments and it also suffers from absences of goal oriented motion. This thesis aims autonomous mobile robot navigation in dynamic environments.

\section{Thesis Objectives}

The main objective of this thesis is autonomous mobile robot navigation in dynamic environments. This is achieved by modifying the DroNet approach proposed in \cite{loquercio2018dronet} to navigate in an indoor environment using ground robot, then retraining the CNN to enhance the performance of DroNet in this environment. In addition to this, we generate path to target so we need a map and a path planning technique. For the map we use simultaneous localization and mapping (SLAM) (gmapping) and for path planning we used Dijkstra. Finally, the combination between the modified DroNet and the paths generated to get a goal oriented motion with shortest path and dynamic obstacle avoidance with low cost. We have implemented and tested their method on both ROS simulation environment and GAZEBO simulation for robotic system. 

\section{Thesis Structure}

This thesis is organized as follows:
Chapter~\ref{ch:intro} presents the thesis motivation, definition of autonomous navigation, the problem statement, the thesis objectives and the thesis structure. In Chapter~\ref{ch:Literature-Review}, a literature review of deep learning-based schemes are studied. Chapter~\ref{ch:CNN} introduces the convolutional neural network (CNN). After that, Chapter~\ref{ch:preparation-studies} studies the simulation tools that we used in our graduation project such as Linux, robot operating system (ROS), C++, python and GAZEBO simulator. Chapter~\ref{ch:proposed_approach} presents our proposed approach and its simulation results. Finally, Chapter~\ref{ch:conclusions} includes the conclusions and future work direction.

%% file: chapters/literature-review.tex
\chapter{Literature Review} 
\label{ch:Literature-Review}

\section{Virtual-to-Real Deep Reinforcement Learning: Continuous Control of Mobile Robots for Mapless Navigation} 

\subsection{Approach}
The approach \cite{tai2017virtual}, presents a learning-based mapless motion planner by taking the sparse 10-dimensional range findings and the target position with respect to the mobile robot coordinate frame as input and the continuous steering commands as output. Traditional motion planners for mobile ground robots with a laser range sensor mostly depend on the obstacle map of the navigation environment where both the highly precise laser sensor and the obstacle map building work of the environment are indispensable. We show that, through an asynchronous deep reinforcement learning method, a mapless motion planner can be trained end-to-end without any manually designed features and prior demonstrations. The trained planner can be directly applied in unseen virtual and real environments. The experiments show that the proposed mapless motion planner can navigate the nonholonomic mobile robot to the desired targets without colliding with any obstacles.

\subsection{Conclusion}
In this approach, a mapless motion planner was trained end-to-end through continuous control deep rienforcement learning (RL) from scratch.We revised the state-of-art continuous deep-RL method so that the training and sample collection can be executed in parallel. By taking the 10-dimensional sparse range findings and the target position relative to the mobile robot coordinate frame as input, the proposed motion planner can be directly applied in unseen real environments without ne-tuning,even though it is only trained in a virtual environment. When compared to the low-dimensional map-based motion planner, our approach proved to be more robust to extremely complicated environments.

\section{GOSELO: Goal-Directed Obstacle and Self-Location Map for Robot Navigation Using Reactive Neural Networks}
\subsection{Approach}
\begin{figure}[H]
	\centering
	\includegraphics[scale=0.6]{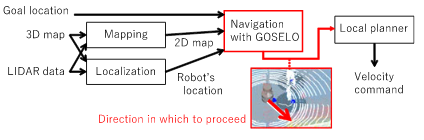}
	\caption{Goselo \cite{kanezaki2017goselo}.}
	\label{fig:goselo}
\end{figure}

Robot navigation using deep neural networks has been drawing a great deal of attention. Although reactive neural networks easily learn expert behaviors and are computationally efficient, they suffer from generalization of policies learned in specific environments. As such, reinforcement learning and value iteration approaches for learning generalized policies have been proposed.However, these approaches are more costly. In the approach \cite{kanezaki2017goselo}, they tackle the problem of learning reactive neural networks that are applicable to general environments. The key concept is to crop, rotate,and resize an obstacle map according to the goal location and the agent’s current location so that the map representation will be better correlated with self-movement in the general navigation task, rather than the layout of the environment. Furthermore, in addition to the obstacle map, we input a map of visited locations that contains the movement history of the agent, in order to avoid failures that the agent travels back and forth repeatedly over the same location as shown in Figure \ref{fig:goselo}. Experimental results reveal that the proposed network outperforms the state-of-the-art value iteration network in the grid-world navigation task. they also demonstrate that the proposed model can be well generalized to unseen obstacles and unknown terrain. Finally, they demonstrate that the proposed system enables a mobile robot to successfully navigate in a real dynamic environment.

The proposed method is based on a convolutional neural network (CNN) to estimate the next best step among neighboring pixels in a grid map. We refer to such a CNN as a “reac-tive CNN” because it reacts to specific patterns on a map in order to determine the movement of the agent. Navigation based on a reactive CNN has three main advantages, as described below.First, a reactive CNN estimates the next best step in a constant time in any situation. In contrast, the computational time of most existing path planning methods, such as the $A^{*}$ search  and rapidly exploring random tree (RRT) depends on the scale and complexity of the map. Furthermore, such classical path planning methods will fail when there is no path to the goal. A CNN-based method can suggest a plausible direction in which to proceed at every moment, regardless of the existence of a path, which is important for navigation in cluttered, dynamic environments. Second, a reactive CNN can use graphics processing unit (GPU) acceleration due to its high potential for parallelization. This is also a major advantage over many classical path planning methods that cannot be wholly parallelized, because every point on a path is dependent on other locations. Finally, a reactive CNN can efficiently learn expert behaviors, e.g., human controls, without modeling the rewards and the policy behind the behaviors.
\subsection{Conclusion}
They proposed a novel navigation method for an online-editable 2D map via an image classification technique. The computation time required by the proposed method to estimate the best direction for the agent remains constant at each step. Another significant advantage of the proposed method is that the agent preferentially moves to new locations, which helps the agent to avoid the local minima trap. Experimental results demonstrated the effectiveness of the proposed goal-directed map representation, i.e., GOSELO, as well as its superiority to existing neural-network-based methods (such as the VIN method) in terms of both success rate and computational cost. We also demonstrated that the proposed method can be generalized to avoid unseen obstacles and navigate unknown terrain. Experiments using the Peacock mobile robot demonstrated the robustness of the proposed navigation system with respect to dynamic scenarios involving crowds of people. Pea-cock successfully moved continuously, all day long for two days, while avoiding people. Peacock demonstrated the advantage of the proposed method over classical path planning methods, such as $A^{*}$ search, which fails to predict the next step when there is no path to the goal. Although we used a CPU for the prediction of a single future step, there would be more room for predicting dozens of future steps if we use a GPU. We are planning such an extension to predict a more reliable direction to proceed. Extending GOSELO from 2D to 3D is another area for future study.

\section{End-to-End Deep Learning for Autonomous Navigation of Mobile Robot}

\subsection{Approach}
\begin{figure}[H]
	\centering
	\includegraphics[scale=0.55]{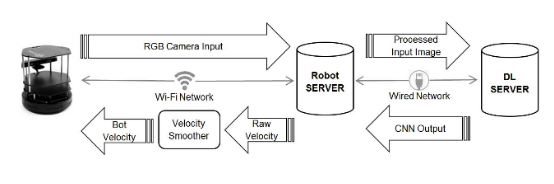}
	\caption{End-to-end deep architecture \cite{kim2018end}.}
	\label{End-to-End Arch.}
\end{figure}

This paper \cite{kim2018end} proposes an end-to-end method for train-ing convolutional neural networks for autonomous navigation of a mobile robot. Traditional approach for robot navigation consists of three steps. The first step is extracting visual features from the scene using the camera input. The second step is to figure out the current position by using a classifier on the extracted visual features. The last step is making a rule for moving the direction manually or training a model to handle the direction.
In contrast to the traditional multi-step method, the proposed visuo-motor navigation system can directly output the linear and angular velocities of the robot from an input image in a singlestep. The trained model gives wheel velocities for navigation asoutputs in real-time making it possible to be implanted on mobilerobots such as robotic vacuum cleaners. The experimental results show an average linear velocity error of 2.2 cm/s and average angular velocity error of 3.03 degree/s. The robot deployed with the proposed model can navigate in a real-world environment by only using the camera without relying on any other sensors such as LiDAR, Radar, IR, GPS, IMU.The proposed system architecture is shown in Figure \ref{End-to-End Arch.}. 

The input of the proposed architecture is an red green blue (RGB) image and the outputs are linear and angular velocities. The system does not require detection, localization or planning modules for navigation separately. The CNN architecture used in this paper is AlexNet . Even though other well-known architectures such as VGGNet , GoogleNet or ResNet can be used, these networks are not applicable for real-time robot navigation due to slow inference speed. Our network performs multi-label regression giving outputs as two real-values. The ground truth velocities are in the range of 0 to 0.5m/s for linear velocity, and in the range of -1.5 to 1.5 radians for angular velocity. Two velocities were normalized to the values between0 and 1 for CNN input. Since the output values are oscillated,using raw output values makes the robot movement unstable. For acquiring consistent outputs, post-processing for noise reduction was conducted to make the movement of the robot stable.

\subsection{Conclusion}
Traditional methods for robot navigation or path planning require multiple and complex algorithms for localization,navigation and action planning. The proposed approach using end-to-end deep learning could make it possible to control the robot motor directly from the visual input as the humandid. The human can decide the path seeing only a local scene without any information of the global map. This result verified the potential of the proposed system as a local path planner.For future work, the visuo-motor system as a global pathplanner can be developed. Moreover, the model can be com-pressed for direct deployment of the visuo-motor system on the embedded board without a server.

\section{From Perception to Decision: A Data-driven Approach to End-to-end Motion Planning for Autonomous
Ground Robots}

\subsection{Approach}

This paper \cite{pfeiffer2017perception} represent a model that is able to learn the complex mapping from
raw 2D-laser range finder and a target position to produce the required steering
commands for the robot. A data-driven end-to-end motion planner based on CNN
model is proposed. The supervised training data is based on expert demonstration
generated using an existing motion planner. The system can navigate the robot
safely though cluttered environment to reach the goal.

Their proposed solution does not require any global map for the robot to navigate.
Given the sensor data and relative target position, the robot is able to navigate
\begin{figure}[H]
	\centering
	\includegraphics[scale=0.55]{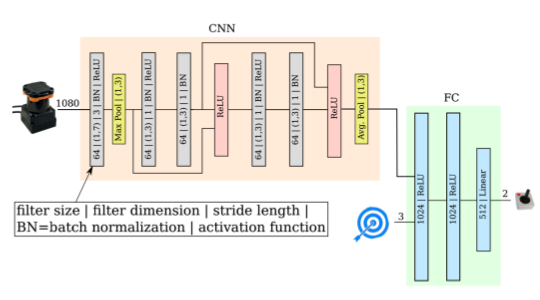}
	\caption{DNN architecture.}
	\label{}
\end{figure}
to the desired location while avoiding the surrounding obstacles. By design, the
approach is not limited to any kind of environment. However, in this paper, only
navigation in static environments is considered. Their main contribution can be
summarized in two points. First, a data-driven end-to-end motion planner from
laser range findings to motion commands. Second, deployment and tests on a
real robotic platform in unknown environments.
The end-to-end relationship between input data and steering commands can
result in an arbitrarily complex model. Among different machine learning ap-
proaches, DNNs/CNNs are well known for their capabilities as a hyper-parametric
function approximation to model complex and highly nonlinear dependencies. To
avoid the problem of generating data for training, simulation data is collected
where a global motion planner is used as an expert. Since no pre-processing of
the laser data is required, the computational complexity and therefore also the
query time for a steering command only depends on the complexity of the model,
which is constant once it is trained.
The paper also consider the case when a robot is faced with a sudden appearing
object blocking the path. Their proposed deep planner reacted clearly to the
object by swerving to the right and after removing the obstacle, it corrects its
course of motion again to reach the target as fast as possible.

\subsection{Conclusion}

This work showed some limitation with real world experimentation. They
found that the robot has some weaknesses when it comes to wide open spaces
with clutter around. However, they suggested that this can potentially results
from the fact that the model was trained purely from perfect simulation data
and re-training using real sensor data might reduce this undesirable effect. In
addition, once the robot enters a convex dead-end region, it is not capable of
freeing itself. Moreover, the motion of the robot sometimes fluctuate apparently
because their architecture does not contain any memory of past visited locations.
Another drawback of the work is that the motion considered are not fully
autonomous and needs sometimes help from a human with a joystick when the
robot got stuck.

%% file: chapters/CNN.tex
\chapter{Convolutional Neural Networks}
\label{ch:CNN}


\section{Artificial Neural Networks}
Even though computers are designed by and for humans, it is clear that the concept of a computer is very different from a human brain. The human brain is a complex and non-linear system, and on top of that its way of processing information is highly parallelized. It is based on structural components known as neurons, which are all designed to perform certain types of computations. It can be applied to a huge amount of recognition tasks, and usually performs these within 100 to 200 ms. Tasks of this kind are still very difficult to process, and just a few years ago, performing these computations on a CPU could take
days \cite{haykin2004comprehensive}. 

Inspired by this amazing system, in order to make computers more suitable for these kinds of task a new way of handling these problems arose. It is called an Artificial neural network (ANN). An ANN is a model based on a potentially massive interconnected network of processing units, suitably called neurons. In order for the network and its neurons to know how to handle incoming information, the model has to acquire knowledge. This is done through a learning process. The connections between the neurons in the network are represented by weights, and these weights store the knowledge learned by the model. This kind of structure results in high generalization, and the fact that the way the neurons handle data can be non-linear is beneficial for a whole range of different applications. This opens up completely new approaches for input-output mapping and enables the creation of highly adaptive models for computation \cite{haykin2004comprehensive}. The learning process itself generally becomes a case of what is called supervised learning, which is described in the next segment.

\section{What is CNN?}
\label{sec:cnn}
\textit{Convolutional Neural Network (CNN, or ConvNet)} is a type of artificial neural network inspired by biological processes \cite{lecun2013deep}. In machine learning, it is a class of deep, feed-forward artificial neural networks that has successfully been applied to analyzing visual imagery. It can be seen as a variant of Multilayer Perceptron (MLP). In computer vision, a traditional MLP connects each hidden neuron with every pixel in the input image trying to find global patterns. However, such a connectivity is not efficient because pixels distant to each other are often less correlated. The found patterns are thus less discriminative to be fed to a classifier. In addition, due to this dense connectivity, the size of parameters grows largely as the size of an input image increases, resulting in substantial increases in both computational complexity and memory space usage. 

However, these problems can be alleviated in CNNs. A hidden neuron in CNNs only connects to a local patch in the input image. This type of sparse connectivity is more effective to discover local patterns and these local patterns learned from one part of an image are also applicable to other parts of the image.

CNNs have been widely used for visual based classification applications. In recent years, a series of R-CNN methods are proposed to apply CNNs on object detection tasks \cite{girshick2014rich, ren2015faster, girshick2015fast, mohamed2019detection}. In \cite{girshick2014rich}, the original version of R-CNN, R-CNN takes full image and object proposals as input. The regional object proposals could come from a variety of methods and in their work they use Selective Search \cite{uijlings2013selective}. Each proposed region is then cropped from the original image and wrapped to a unified $227 \times 227$ pixel size. A 4096-dimensional feature vector is extracted by forward propagating the subtracted region through fine-tuned CNN with \textit{five convolutional layers} and \textit{two fully connected layers}. With the feature vectors, a set of class-specific linear support vector machines (\textit{SVMs}) are trained for classifications.

R-CNN achieves excellent object detection accuracy, however, it has notable drawbacks. First, training and testing has multiple stages including fine-tuning CNN with \textit{Softmax} loss, training SVMs and learning bounding-box regressors. Secondly, the CNN part is slow because it performs forward pass for each object proposal without sharing computation. To address the speed problem, Spatial Pyramid Pooling network (SPPnet) \cite{he2014spatial} and Fast R-CNN \cite{girshick2015fast} are proposed. Both methods compute one single convolutional feature map for the entire input image and do the cropping on the feature map instead of on the original image and then extract feature vectors for each region. For feature extraction, SPPnet pools the feature maps into multiple sizes and concatenate them as a spatial pyramid \cite{lazebnik2006beyond}, while Fast R-CNN only use single scale of the feature maps. The feature sharing of SPPnet accelerates R-CNN by 10 to 100x in testing and 3x in training. However it still has the same multiple-stage pipeline as R-CNN. In Fast R-CNN Girshick propose a new type of layer, region of interest (RoI) pooling layer, to connect the gap between feature maps and classifiers. With this layer, they build an \textit{semi} end-to-end training framework which only rely on full image input and object proposals.

All the mentioned methods rely on external object proposal input. In \cite{ren2015faster}, the authors proposed proposal-free framework called Faster R-CNN. In Faster R-CNN, they use a RPN, which slides over the last convolutional feature maps to generate bounding-box proposals in different scales and ratio aspects. These proposals are then fed back to Fast R-CNN as input. Another proposal-free work \textit{You Only Look Once} is proposed in \cite{redmon2016you}. This network uses features from the entire image to predict object bounding box. Instead of sliding windows on the last convolutional feature maps, this network connects the feature map output to an 4096-dimensional followed by another full-connected $7\times7\times24$ tensor. The tensor is a $7\times7$ mapping of the input image. Each grid of the tensor is a 24-dimensional vector which encodes bounding boxes and class probabilities of the object whose center falls into this grid on the origin image. The YOLO network is 100 to 500x faster than Fast R-CNN based methods, though with less than 8\% \textit{mean Average Precision} (mAP) drop on VOC 2012 test set \cite{everingham2015pascal}.

Some other specific R-CNN variants are also proposed to solve different problems. The paper from \cite{gkioxari2014r} presents a R-CNN based networks with triple loss functions combined for the task of keypoints (as representation pose) prediction and action classification of people. It also adapt R-CNN to use more than one region, but also contextual subregions for human detection and action classification called R*CNN \cite{gkioxari2015contextual}. In \cite{ouyang2015deepid}, the authors proposed DeepID-Net with deformation constrained pooling layer, which models the deformation of object parts with geometric constraint and penalty. Furthermore, a broad survey of the recent advances in CNNs and its applications in computer vision, speech and natural language processing have been presented in \cite{gu2015recent}. 

On the other hands, in general, there are some effective laser-based methods for object detection, estimation and tracking using machine learning approaches \cite{barbiereal, teichman2011practical, xiaoa2016simultaneous, pinto2013object}. A multi-modal system for detecting, tracking and classifying objects in outdoor environment was presented in \cite{premebida2007lidar}.

\section{Network Structures and Essential Layers}
In this section, some important concepts related to general CNNs including the structure of networks and essential layers will be covered.
 
\subsection{CNN Architectures}
\label{subsec:cnnArch}
\begin{figure}[h]
\centering
\includegraphics[scale=0.38]{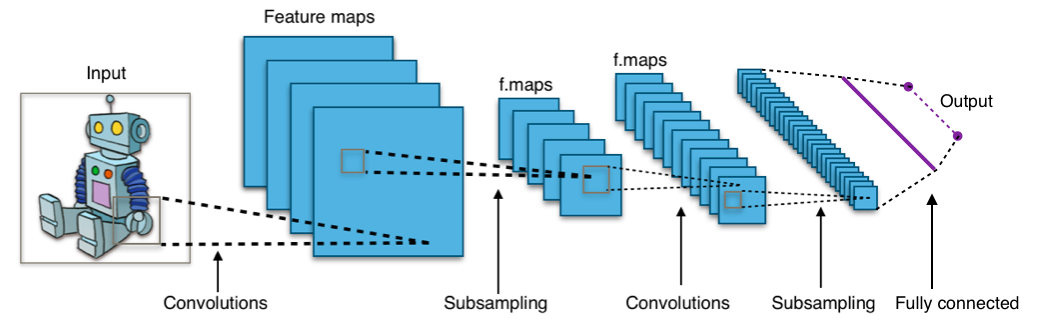}
\caption{A typical architecture of CNN (Wikipedia).}
\label{fig:typicalcnn}
\end{figure}

In general, the architecture of CNN can be decomposed into two stages, which are hierarchical feature extraction stage and classification stage. A typical architecture of CNN is shown in Figure \ref{fig:typicalcnn}. An input image is convolved by a set of trainable filters (kernels) each with a nonlinear mapping (e.g. ReLU \cite{nair2010rectified}) to produce so-called \textit{feature maps}. Each feature map containing special features is then partitioned into equal-sized, non-overlapping regions and the maximum (or average) of each region is passed to the next layer (sub-sampling layer), resulting in resolution-reduced feature maps with depth unchanged. This operation allows small translation to the input image, thus more robust features that are invariant to translations are more likely to be found \cite{goodfellowdeep}. These two steps, convolutions and subsampling, are alternated for two iterations in the CNN in Figure \ref{fig:typicalcnn} and the resulting feature maps are fully connected with a MLP to perform classification. In some applications, the final fully connected layer that performs classification is replaced with other classifiers e.g. SVM. For example, the state-of-the-art object detector R-CNN \cite{girshick2014rich} extracts high-level features from the penult final fully layer and feeds them to SVMs for classification \cite{braun2016pose}.

\subsection{Layers in CNNs}
\label{subsec:layercnn}
As mentioned in Section \ref{subsec:cnnArch}, CNNs are commonly made up of mainly three layer types: \textit{convolutional layer, pooling layer (usually subsampling) and fully connected layer}. The explanations of these layers and the introduction of other auxiliary layers that are not shown in Figure \ref{fig:typicalcnn} will be introduced. 

\begin{itemize}
\item \textit{Convolution Layer}\vspace{0.3cm}\\
The convolutional layer is the core building block of a CNN. The layer's parameters consist of a set of filters or kernels, which have a small receptive field, but extend through the full depth of the input volume or image. The convolution operation replicates a filter across the entire image field to get the response of each location and form a response feature map. Given multiple filters, the network will get a stack of features maps to form a new 3D volume. 

Officially, the convolution layer accepts a volume or image of size $W_1 \times H_1 \times D_1$ from previous layer as input data, where $H_1 , W_1, D_1$ are image height,  image width, and a number of channels (or depth) respectively. The layer defines $K$ filters with the shape $F \times F \times D_1$ each, where $F$ is the kernel size. The convolution of input volume and filters produces the output volume of size $W_2 \times H_2 \times K$, where the new volume's $W_2$ and $H_2$ are dependent on the filter size, stride and pad settings of the convolution operation. In general, the formula for calculating the output size, $W_2 \text{and} H_2$, for any given convolution layer is defined as:
\begin{itemize}
\item width: $W_2 = \frac{(W_1 - F + 2P )}{S} + 1$,
\item height: $H_2 = \frac{(H_1 - F + 2P )}{S} + 1$,
\end{itemize}
Where: $K$ is the filter size, $P$ is the padding, and $S$ is the stride.
	
For instance, Figure \ref{fig:convol} illustrate a 2D version convolution where the $7 \times 7 \times 1$ input volume is convolved with one $3 \times 3$ filter. With $0$ padding and 1 stride settings, it produces a $5 \times 5 \times 1$ output volume.

\begin{figure}[h]
\centering
\includegraphics[scale=1.3]{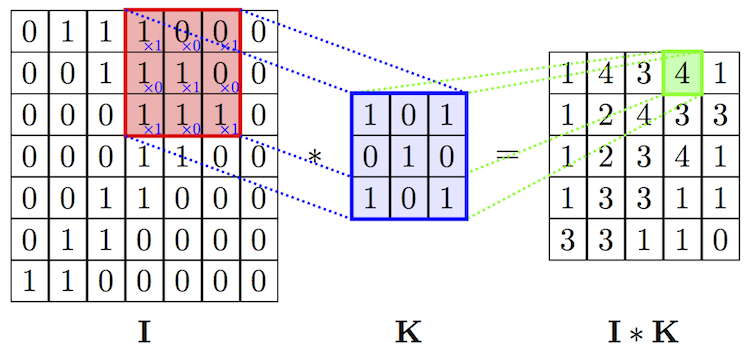}
\caption{An example of convolution operation in 2D \cite{mohamed2017detection}.}
\label{fig:convol}
\end{figure}

\begin{itemize}
\item \textit{Stride and Padding}\\
There are two main parameters that must be tuned after choosing the filter size $K$ in order to modify the behavior of each layer. These two parameters are the \textit{stride} and the \textit{padding}. \textit{Stride}, $S$, controls how the filter convolves around the input volume. In the previous example, $S=1$. This means that the filter convolves around the input volume by shifting one unit at a time. So, the amount by which the filter shifts is the stride. Moreover, as we keep applying convolution layers, the size of the volume will decrease faster than we would like. Therefore, in order to preserve as much information about the original input volume so that we can extract those low level features, the zero-padding must be applied. Let's say we want to apply the same convolution layer but we want the output volume to remain $7 \times 7 \times 1$, which is equal to the input size. To do this, we can apply a zero-padding of size 1 to that layer. Zero-padding pads the input volume with zeros around the border. The zero-padding is defined as:
\begin{equation}
P = \frac{K-1}{2}.
\end{equation}

\end{itemize}
\item \textit{Pooling Layer}\vspace{0.3cm}\\
Another important concept of CNNs is pooling, which is a form of non-linear down-sampling. It partitions the input image into a set of non-overlapping rectangles and, for each such sub-region, outputs the maximum (in case of \textit{max-pooling}). The function of pooling layer is to reduce the spatial size of representation and hence reduce the amount of parameters and amount of computations in the network and also control over-fitting. There are several non-linear functions to implement pooling such as \textit{max-pooling, average-pooling and stochastic-pooling}. The pooling layer operates independently on every depth slice of the input and resizes it spatially. Pooling is an translation-invariance operation. The pooled image keeps the structural layout of the input image.

Formally a pooling layer accepts a volume of size $W_1 \times H_1 \times D_1$ as input and output a volume of size $W_2 \times H_2 \times D_1$. The output width $W_2$ and height $H_2$ are dependent on the kernel size, stride and pad settings, as shown in Figure \ref{fig:maxpool}. The produced output has dimensions:

\begin{itemize}
\item width: $W_2 = \frac{(W_1 - F)}{S} + 1$, and 
\item height: $H_2 = \frac{(H_1 - F)}{S} + 1$.
\end{itemize}

\begin{figure}[h]
\centering
\includegraphics[scale=0.7]{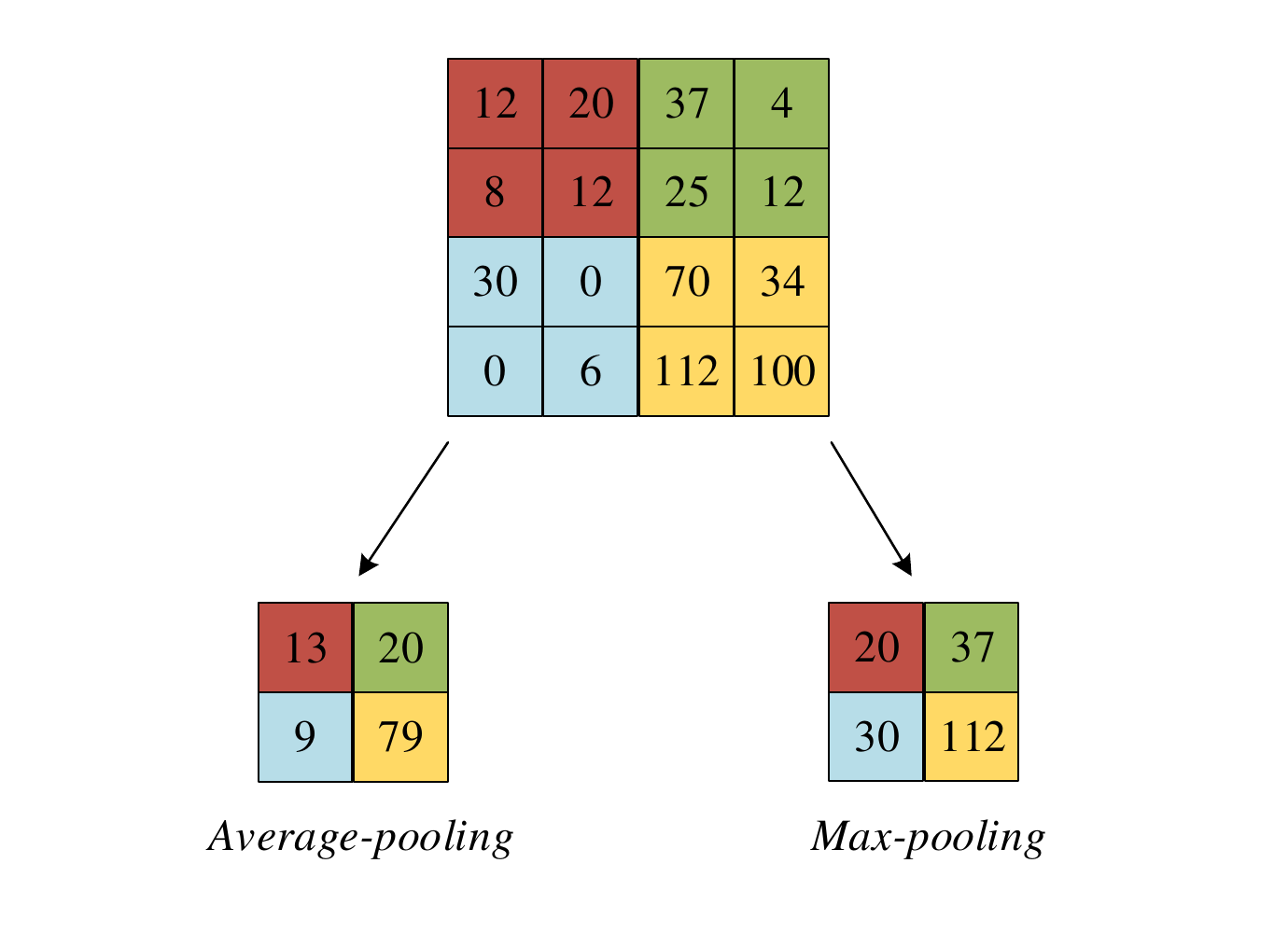}
\vspace{-0.7cm}
\caption{An example of pooling with a $2\times2$ filter and a stride of 2 \cite{mohamed2017detection}.}
\label{fig:maxpool}
\end{figure}

\begin{figure}[h]
\centering
\includegraphics[scale=0.5]{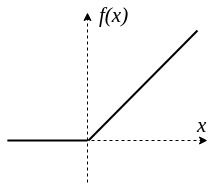}
\vspace{-0.4cm}
\caption{The ReLU activation function.}
\label{fig:relu}
\end{figure}
\item \textit{ReLU Layer}\vspace{0.3cm}\\
Rectified Linear Units (ReLU) is one of the most notable non-saturated activation functions, which can be used by neurons just like any other activation function. The ReLU activation function is defined as (Figure \ref{fig:relu}):
\begin{equation}
f(x) = \text{max}(0, x).
\end{equation}
ReLU is an element wise operation (applied per pixel) and replaces all negative pixel values in the feature map by zero. It increases the nonlinear properties of the decision function and of the overall network without affecting the receptive fields of the convolution layer. For that reason, after each convolution layer, it is convention to apply a ReLU layer immediately afterwards. The main reason that it is used is because of how efficiently it can be computed compared to more conventional activation functions like the \textit{sigmoid function} $f(x) = \vert tanh(x) \vert$ and \textit{hyperbolic tangent function} $f(x) = tanh(x)$, without making a significant difference to generalization accuracy. Many works have shown that ReLU works better than other activation functions empirically \cite{maas2013rectifier, he2015delving}. Moreover, the recently used activation functions in CNNs based on ReLU such as Leaky ReLU \cite{maas2013rectifier}, Parametric ReLU \cite{he2015delving}, Randomized ReLU \cite{xu2015empirical}, and Exponential Linear Unit (ELU) \cite{clevert2015fast} were introduced in \cite{gu2015recent}.
  
\item \textit{Fully Connected Layer}\vspace{0.3cm}\\
Eventually, after several convolutional and max pooling layers, the high-level reasoning in the neural network is done via fully connected layers. Neurons in a fully connected layer have full connections to all neurons in the previous layer. It provides a form of dense connectivity and loses the structural layout of the input image. Fully connected layers are usually inserted after the last convolution layer to reduce the amount of features and creating vector-like representation.

\item \textit{Loss Layer}\vspace{0.3cm}\\
It is important to choose an appropriate loss function for a specific task. The loss layer specifies the learning process by comparing the output of the network with the true label (or target) and minimizing the cost. Generally, the loss is calculated by forward pass and the gradient of network parameters with respect to loss is calculated by the backpropagation. For multi-class classification problems, softmax classifier with loss is commonly used. Firstly, it takes multi-class scores as input, and uses softmax function to normalize the input and get a distribution-like output. Then, the loss is computed by calculating the cross-entropy of the target class probability distribution and the estimated distribution. The softmax function is defined as:
\begin{equation}
y(x)_i = \frac{\text{exp}(x_i)}{\sum_{j=1}^{n}\text{exp}(x_j)},
\end{equation}
Where:
\begin{itemize}
\item  $0 \leq y(x)_i \leq 1$,
\item  $\sum_{j=1}^{n}y(x)_j =1$,
\item  $i = 1,\ldots, n$ \& $ n$ is the number of Classes.
\end{itemize}
 The cross-entropy between the target distribution $p$ and the estimation distribution  $q$ is given by
 \begin{equation}
 H(p,q) = \sum_i p_i \text{log}q_i.
 \end{equation}

The purpose of the softmax classification layer is simply to transform all the network activations in your final output layer to a series of values that can be interpreted as probabilities. The softmax function is also known as the normalized exponential function. The recently used loss layers (e.g. Hinge loss \cite{zhang2004solving}, L-Softmax loss \cite{liu2016large}, Contrastive loss \cite{chopra2005learning, hadsell2006dimensionality}, and Triplet loss \cite{schroff2015facenet}), were presented in \cite{gu2015recent}.
\end{itemize}


%% file: chapters/Preparation-Studies.tex
\chapter{Preparation Studies} 
\label{ch:preparation-studies}
\section{Mobile Robots}
Mobile robots are vehicles with the ability to change their positions. These robots can move on grounds, on the surface of water, under water and in the air. Two modes can be used to operate mobile robots. One is tele-operated mode where movement instructions are given externally. Another mode is autonomous where robots operate on the information that these get from sensors and no external instructions are given. Wheeled mobile robots are one of the types of mobile robots extensively used in research and industry, as wheel is the most popular locomotion mechanism in mobile robotics. One of the advantages of wheeled robots is that balancing is not a problem as robots are designed in such a way that all wheels are on the ground. The Figure Below show examples of mobile robots.
\begin{figure}[H]
	\centering
	\includegraphics[scale=0.2]{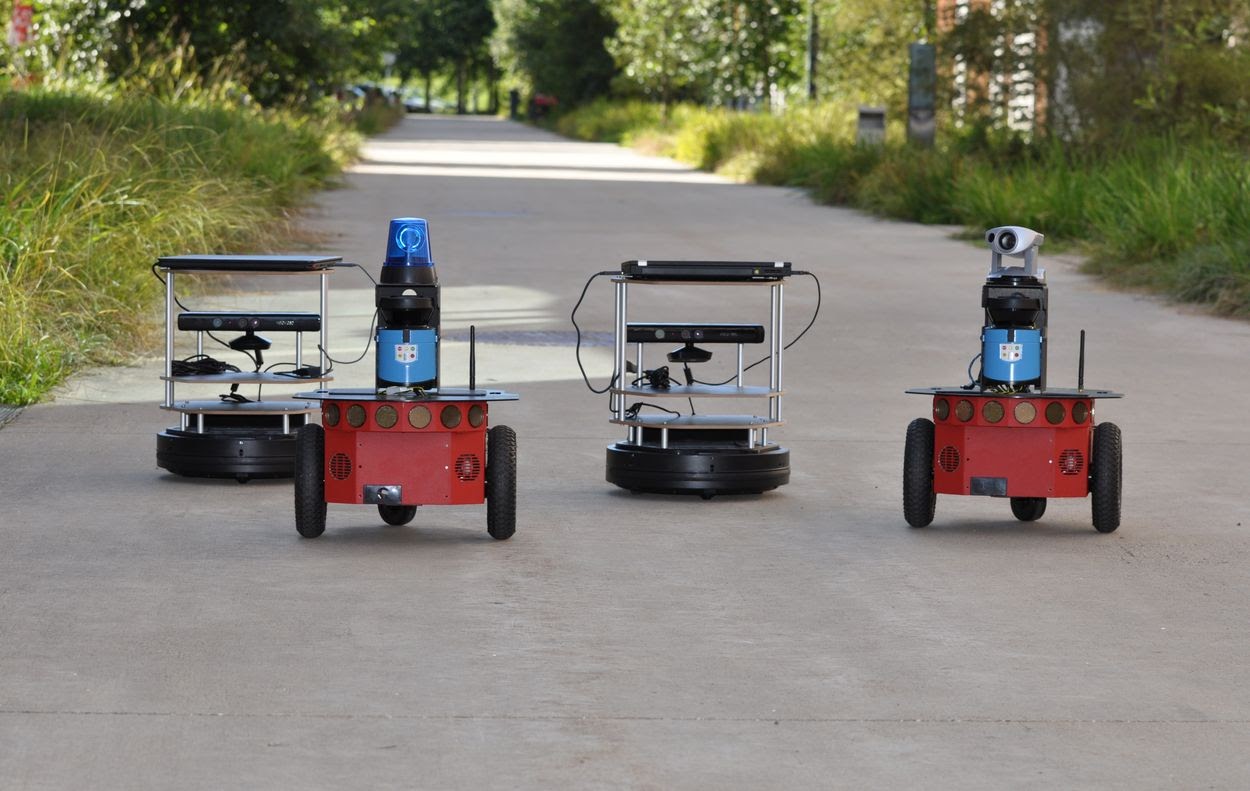}
	\caption{Mobile robots\protect\footnotemark{}.}
	\label{fig:mobile robots examples}
\end{figure}
\footnotetext{\url{https://robohub.org/robot-teams-create-supply-chain-to-deliver-energy-to-explorer-robots/}}
\section{Kinematics of Differential Drive Robots}
\begin{figure}[H]
	\centering
	\includegraphics[scale=0.75]{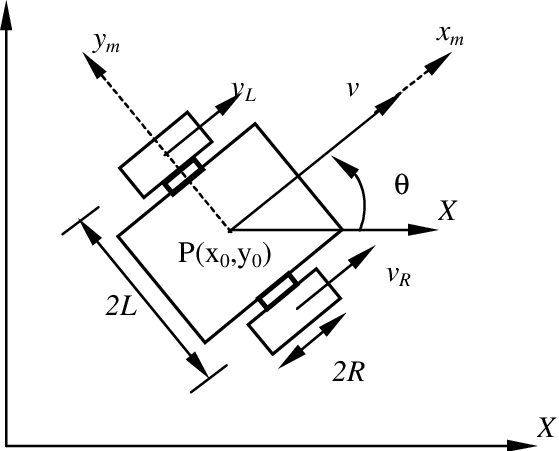}
	\caption{Differential robot model \cite{rashid2014simulation}.}
	\label{fig:differential robot model}
\end{figure}

The differential drive robot is, probably, the most common and most used mobile robot in the current times. A differential drive robot consists of two independently driven wheels that rotate about the same axis, as well as one or more caster wheels, ball casters, or low-friction sliders that keep the robot horizontal. This is the case with the robot in our simulation. Figure \ref{fig:differential robot model} shows a visual representation of the system.
For the differential drive system, you are going to need to know 2 parameters:
\begin{itemize}
  \item L: The distance between the wheels of the robot, also known as Wheel Base.
  \item R: The radius of the wheels of the robot.
\end{itemize}

These parameters are relatively easy to measure in any system. On a real robot, can just measure them with a ruler, On a simulated robot, you can also extract these values from the unified robot deccription format (URDF) file of the robot.

From the visual representation above, the 2 inputs for the system:
\begin{itemize}
  \item $v_R$: Rate at which the right wheel is turning
  \item $v_L$: Rate at which the left wheel is turning
\end{itemize}

So, in order to have the kinematic model of our system, a set of equations that connect the inputs of our system with the outputs are required. For the differential drive robot, these are the equations:

\begin{equation}
\begin{array}{l}
\dot{x}=\frac{R}{2}\left(v_{r}+v_{l}\right) \cos (\theta) \\
\dot{y}=\frac{R}{2}\left(v_{r}+v_{l}\right) \sin (\theta) \\
\dot{\theta}=\frac{R}{L}\left(v_{r}-v_{l}\right)
\end{array}
\end{equation}

\section{The Purpose of Using Robot Operating System (ROS)}
\begin{figure}[H]
	\centering
	\includegraphics[scale=0.43]{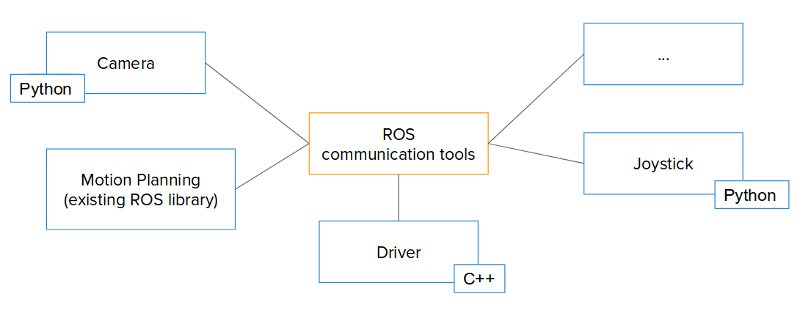}
	\caption{ROS with libraries \cite{dahl1972structured}\protect\footnotemark{}.}
	\label{fig:ros with libraries}
\end{figure}

Robot Operating System (ROS) allows you to stop reinventing the wheel. Reinventing the wheel is one of the main killers for new innovative applications. The ROS goal is to provide a standard for robotics software development, that you can use on any robot. Whether you are programming a mobile robot, a robotic arm, a drone, a boat, a vending machine, You can use the Robot Operating System. This standard allows you to actually focus on the key features of your application, using an existing foundation, instead of trying to do everything yourself. ROS is more of a middle ware, something like a low-level “framework” based on an existing operating system. The main supported operating system for ROS is Ubuntu. You have to install ROS on your operating system in order to use it. Robot Operating System is mainly composed of 2 things:

\begin{itemize}
  \item a core (middle ware) with communication tools,
  \item a set of plug \& play libraries.
\end{itemize}
Basically, a middle ware is responsible for handling the communication between programs in a distributed system (as shown in figure \ref{fig:ros with libraries}) 

\footnotetext{\url{https://roboticsbackend.com/what-is-ros/}}

ROS comes with 3 main communication tools:
\begin{itemize}
  \item Topics: Those will be used mainly for sending data streams between nodes. Example: you’re monitoring the temperature of a motor on the robot. The node monitoring this motor will send a data stream with the temperature. Now, any other node can subscribe to this topic and get the data.
  \item Services: They will allow you to create a simple synchronous client/server communication between nodes. Very useful for changing a setting on your robot, or ask for a specific action: enable freedrive mode, ask for specific data, etc.
  \item Actions: A little bit more complex, they are in fact based on topics. They exist to provide you with an asynchronous client/server architecture, where the client can send a request that takes a long time (ex: asking to move the robot to a new location). The client can asynchronously monitor the state of the server, and cancel the request anytime.
  
\end{itemize}

\section{Sensors}
Robots must sense the world around them in    order    to react to variations in    tasks and environments. The sensors can range from minimalist setups designed for quick installation to highly elaborate and tremendously expensive sensor rigs.
Many successful industrial deployments     use surprisingly little sensing. A remarkable number of    complex and intricate industrial manipulation tasks can    be performed through    a combination of    clever mechanical engineering and    limit    switches, which close or open an electrical circuit when a mechanical    lever    or plunger    is pressed,     in order to start execution of    a pre-programmed    robotic manipulation sequence. Through careful mechanical setup and tuning,    these systems can achieve    amazing levels of throughput     and reliability. It is important,    then,    to consider    these    binary sensors when enumerating    the world of robotic sensing. These sensors are typically either “on” or “off.” In addition to    mechanical limit switches, other binary sensors include optical limit switches, which use a mechanical  “flag” to interrupt a light beam, and bump sensors, which channel mechanical pressure along a relatively large distance    to a single    mechanical switch. These relatively simple sensors    are a key part of    modern industrial automation equipment, and their importance can hardly be overstated.
Another class of sensors return scalar readings. For example, a pressure sensor can estimate the mechanical or barometric pressure and will typically output a scalar value along some range of sensitivity chosen at time of manufacture. Range sensors can be constructed from many physical phenomena (sound, light, etc.) and will also typically return a scalar value in some range, which seldom includes zero or infinity!
Each sensor class has its own quirks that distort its    view of reality and must be
accommodated by sensor-processing algorithms. These quirks can often be surprisingly severe. For example,    a range sensor may have a “minimum distance” restriction: if an object is closer than that minimum distance, it will not    be sensed. As a result of these quirks, it is often advantageous to combine several different types of sensors in a robotic system.

\subsection{Visual Cameras}
Higher-order animals tends to rely on visual data to react    to the world around    them. If only robots were as    smart    as animals! Unfortunately, using camera    data    intelligently is surprisingly difficult, as we will describe in later chapters of this book. However,    cameras are cheap and often useful for tele-operation, so it is    common to see them on robot sensor heads.
Interestingly, it is often more mathematically robust to describe robot tasks and environments in three dimensions    (3D) than it is to work with 2D camera images. This is because the 3D shapes of tasks and environments are invariant to changes in scene lighting, shadows, occlusions, and so on. In fact, in  a surprising number of application domains, the visual data is largely ignored; the algorithms are interested in 3D data. As a result, intense research efforts have been expended on producing 3D data of the scene in front of the robot.
When two cameras are rigidly mounted to a common mechanical structure, they form    a stereo camera. Each camera sees a slightly different view of the world, and these slight differences can be used to estimate the distances to    various features in    the image. This sounds simple, but as always, the devil is in the details. The performance of a stereo camera depends on a large number of factors, such as the quality of the camera’s mechanical design,    its resolution, its    lens type and quality, and so on. Equally important are the qualities of the scene being imaged: a    stereo camera can only estimate the distances to mathematically discernible features in the scene, such as sharp, high-contrast
Corners. A stereo camera cannot, for example, estimate the distance to a featureless wall, although it can most likely estimate the distance to the corners and edges of the wall, if they intersect a floor, ceiling, or other wall of a different color. Many natural outdoor scenes possess sufficient texture that stereo vision can be made to work quite well for depth estimation. Uncluttered indoor scenes, however, can often be quite difficult.
Several conventions have emerged in the ROS community for handling cameras. The canonical ROS message type for images is sensor\_msgs/Image , and it contains little more than the size of the image , its pixel encoding scheme, and the pixels themselves. To describe the intrinsic distortion of the camera resulting from its lens and sensor alignment, the  sensor\_msgs/CameraInfo message is used. Often, these ROS images need to be sent to    and from OpenCV, a popular computer vision library.

\subsection{An Image Processing Example of ROS Architecture}
\begin{figure}[H]
	\centering
	\includegraphics[scale=0.4]{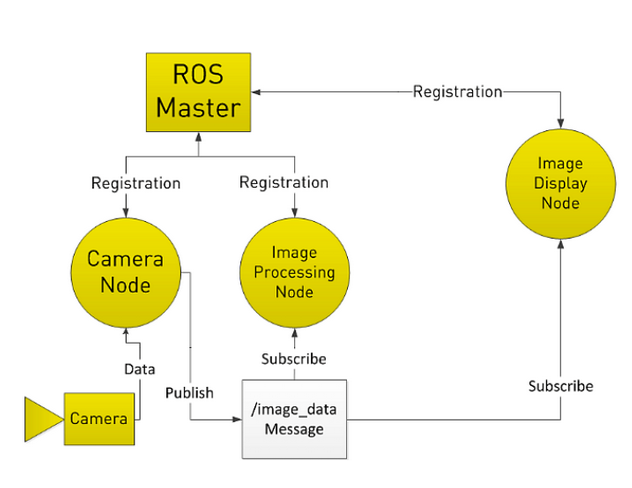}
	\caption{An image processing example of ROS architecture\protect\footnotemark{}.}
	\label{Visual camera diagram.}
\end{figure}
\footnotetext{\url{https://www.researchgate.net/figure/An-image-processing-example-of-ROS-architecture-The-Camera-Node-publishes-images-in-a_fig3_341992473}}

The Camera Node publishes images in a message named image\_data which is subscribed by both Image Display Node and Image Processing Node. The ROS Master tracks publishers and subscribers enabling individual nodes to locate and message each other.

\subsection{Depth Cameras}
\begin{figure}[H]
	\centering
	\includegraphics[scale=0.5]{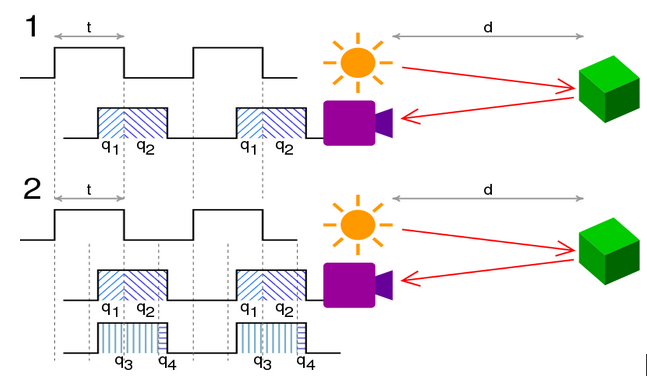}
	\caption{Principle of operation of a time-of-flight camera\protect\footnotemark{}.}
	\label{Principle of operation of a time-of-flight camera.}
\end{figure}

As discussed in the previous section, even though visual camera data is intuitively appealing, and seems like it should be useful somehow, many perception algorithms work much better with 3D data. Fortunately, the past few years have seen massive progress in low-cost depth cameras. Unlike the passive stereo cameras described in the previous section, depth cameras are active devices. They illuminate the scene in various  ways, which greatly improves the system performance. For example, a completely featureless indoor wall or surface is essentially impossible to detect using passive stereo vision.
However, many depth cameras will shine a texture pattern on the surface, which is subsequently imaged by its camera. The texture pattern and camera are typically set to operate in near-infrared wavelengths to reduce the system’s sensitivity to the colors of objects, as well as to not be distracting to people nearby.
\begin{figure}[H]
	\centering
	\includegraphics[scale=0.55]{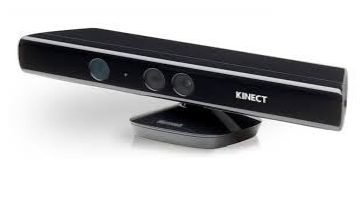}
	\caption{Depth camera.}
	\label{Depth camera.}
\end{figure}
Some common depth cameras, such as the Microsoft Kinect as shown in Figure \ref{Depth camera.}, project a structured light image. The device projects a precisely known pattern into the scene, its camera observes how this pattern is deformed as it lands on the various objects and surfaces of the scene, and finally a reconstruction algorithm estimates the 3D structure of the scene from this data. It’s hard to overstate the impact that the Kinect has had on modern robotics! It was designed for the gaming market, which is orders of magnitude larger than the robotics sensor market, and could justify massive expenditures for the development and production of the sensor. The launch price of 150\$ was incredibly cheap for a sensor capable of outputting so much useful data. Many robots were quickly retrofitted to hold Kinects, and the sensor continues to be used across research and industry. Although the Kinect is the most famous (and certainly the most widely used) depth camera in robotics, many other depth-sensing schemes are possible. For example, unstructured light depth cameras employ “standard” stereo-vision algorithms with random texture injected into the scene by some sort of projector. This scheme has been shown to work far better than passive stereo systems in feature-scarce environments, such as many indoor scenes.
A different approach is used by time-of-flight depth cameras. These imagers rapidly blink an infrared light emitting diode (LED) or laser illuminator, while using specially designed pixel structures in their image sensors to estimate the time required for these light pulses to fly into the scene and bounce back to the depth camera. Once this “time of flight” is estimated, the (constant) speed of light can be used to convert the estimates into a depth image, as illustrated in Figure \ref{Principle of operation of a time-of-flight camera.} .
\footnotetext{\url{https://en.wikipedia.org/wiki/Time-of-flight\_camera\#/media/File:Time\_of\_flight\_camera\_principle.svg}}
Intense research and development is occurring in this domain, due to the enormous existing and potential markets for depth cameras in video games and other mass-market user-interaction scenarios. It is not yet clear which (if any) of the schemes discussed previously will end up being best suited for robotics applications. At the time of writing, cameras using all of the previous modalities are in common usage in robotics experiments.
Just like visual cameras, depth cameras produce an enormous amount of data. This data is typically in the form of point clouds, which are the 3D points estimated to lie on the surfaces facing the camera. The fundamental point cloud message is sensor\_msgs/PointCloud2 (so named purely for historical reasons). This message allows for unstructured point cloud data, which is often advantageous, since depth cameras often cannot return valid depth estimates for each pixel in their images. As such, depth images often have substantial “holes,” which processing algorithms must handle gracefully.
\subsection{Laser Scanners}
\begin{figure}[H]
	\centering
	\includegraphics[scale=0.5]{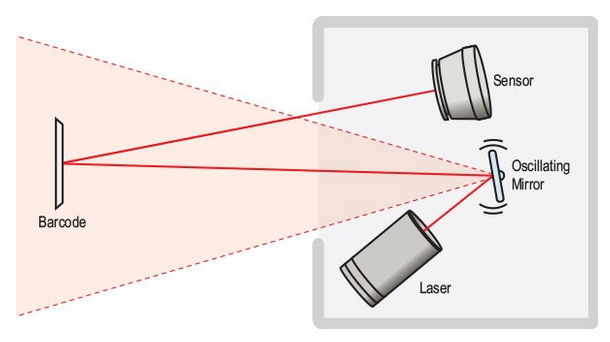}
	\caption{Laser scanner diagram\protect\footnotemark{}.}
	\label{Laser scanner diagram.}
\end{figure}

Although depth cameras have greatly changed the depth-sensing market in the last few years due to their simplicity and low cost, there are still some applications in which laser scanners (Figure \ref{Laser scanner diagram.}) are widely used due to their superior accuracy and longer sensing range. There are many types of laser scanners, but one of the most common schemes used in robotics involves shining a laser beam on a rotating mirror spinning around 10 to 80 times per second (typically 600 to 4,800 RPM). As the mirror rotates, the laser light is pulsed rapidly, and the reflected waveforms are correlated with the outgoing waveform to estimate the time of flight of the laser pulse for a series of angles around the scanner.
\footnotetext{\url{https://www.cognex.com/en-nl/what-is/industrial-barcode-reading/laser-scanners}}

Laser scanners used for autonomous vehicles are considerably different from those used for indoor or slow-moving robots. Vehicle laser scanners made by companies such as Velodyne must deal with the significant aerodynamic forces, vibrations, and temperature swings common to the automotive environment. Since vehicles typically move much faster than smaller robots, vehicle sensors must also have considerably longer range so that sufficient reaction time is possible. Additionally, many software tasks for autonomous driving, such as detecting vehicles and obstacles, work much better when multiple laser scan lines are received each time the device rotates, rather than just one. These extra scan lines can be extremely useful when distinguishing between classes of objects, such as between trees and pedestrians. To produce multiple scanlines, automotive laser scanners often have multiple lasers mounted together in a rotating structure, rather than simply rotating a mirror. All of these additional features naturally add to the complexity, weight, size, and thus the cost of the laser scanner.
The complex signal processing steps required to produce range estimates are virtually always handled by the firmware of the laser scanner itself. The devices typically output a vector of ranges several dozen times per second, along with the starting and stopping angles of each measurement vector. In ROS, laser scans are stored in sensor\_msgs/LaserScan messages, which map directly from the output of the laser scanner. Each manufacturer, of course, has their own raw message formats, but ROS drivers exist to translate between the raw output of many popular laser scanner manufacturers and the sensor\_msgs/LaserScan message format.

\section{TurtleBot}
TurtleBot is the robot we used in this thesis, The TurtleBot was designed in 2011 as a minimalist platform for ROS-based mobile robotics education and prototyping. It has a small differential-drive mobile base with an internal battery, power regulators, and charging contacts. Atop this base is a stack of laser- cut “shelves” that provide space to hold a netbook computer and depth camera, and lots of open space for prototyping. To control cost, the TurtleBot relies on a depth camera for range sensing; it does not have a laser scanner. Despite this, mapping and navigation can work quite well for indoor spaces. TurtleBot are available from several manufacturers for less than 2,000\$. More information is available at \url{http://turtlebot.org}.
\begin{figure}[H]
	\centering
	\includegraphics[scale=0.55]{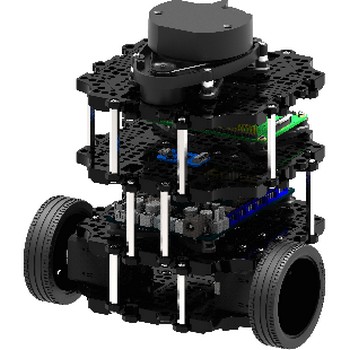}
	\caption{Turtlebot burger\protect\footnotemark{}.}
	\label{TurtleBot.}
\end{figure}
\footnotetext{\url{https://www.robot-advance.com/EN/art-turtlebot3-burger-1997.htm}}
Because the shelves of the TurtleBot (in Figure \ref{TurtleBot.}) are covered with mounting holes, many owners have added additional subsystems to their TurtleBot, such as small manipulator arms, additional sensors, or upgraded computers. However, the “stock” TurtleBot is an excellent starting point for indoor mobile robotics. Many similar systems exist from other vendors, such as the Pioneer and Erratic robots and thousands of custom- built mobile robots around the world. The examples in this book will use the TurtleBot, but any other small differential-drive platform could easily be substituted.

\section{Navigation}
\subsection{ROS Navigation }
ROS has a set of resources that are useful so a robot is able to navigate through a medium, in other words, the robot is capable of planning and following a path while it deviates from obstacles that appear on its path throughout the course. These resources are found on the navigation stack. One of the many resources needed for completing this task and that is present on the navigation stack are the localization systems, that allow a robot to locate itself, whether there is a static map available or simultaneous localization and mapping is required. Adaptive Monte Carlo Localization (AMCL) is a tool that allows the robot to locate itself in an environment through a static map, a previously created map. The disadvantage of this resource is that, because of using a static map, the environment that surrounds the robot can not suffer any modification, because a new map would have to be generated for each modification and this task would consume computational time and effort. Being able to navigate only in modification-free environments is not enough, since the robots should be able to operate in places like industries and schools, where there is constant movement. To bypass the lack of flexibility of static maps, two other localization systems are offered by the navigation stack: gmapping and hector mapping. Both gmapping and hector mapping are based on Simultaneous Localization and Mapping (SLAM), a technique that consists of mapping an environment at the same time that the robot is moving, in other words, while the robot navigates through an environment, it gathers information from the environment through his sensors and generates a map. This way you have a mobile base able not only to generate a map of an unknown environment as well as updating the existing map, thus enabling the use of the device in more generic environments, not immune to changes. The difference between gmapping and hector mapping is that the first one takes in account the odometry information to generate and update the map and the robot’s pose, however, the robot needs to have encoders, preventing some robots(e.g. flying robots) from using it. The odometry information is interesting because they are able to aid in the generation of more precise maps, since understanding the robot dynamics we can estimate its pose. The dynamic behaviour of the robot is also known as kinematics. Kinematics is influenced, basically, by the way that the devices that guarantee the robot’s movement are assembled. Some examples of mechanical features that influence the kinematics are: the wheel type, the number of wheels, the wheels positioning and the angle at which they are disposed. However, as much useful as the odometry information can be, it isn’t immune to faults. The faults are caused by the lack of precision on the capitation, friction, slip, drift and other factors, and, with time, they may accumulate, making inconsistent data and prejudicing the maps formation, that tend to be distorted under these circumstances. Other indispensable data to generate a map are the sensors‘ distance readings, for the reason that they are responsible in detecting the external world and, this way, serve as reference to the robot. Nonetheless, the data gathered by the sensors must be adjusted before being used by the device. These adjustments are needed because the sensors measure the environment in relation to themselves, not in relation to the robot, in other words, a geometric conversion is needed. To make this conversion simpler, ROS offers the TF tool, which makes it possible to adjust the sensor's positions in relation to the robot and, this way, adequate the measures to the robot‘s navigation.

\subsubsection{The Navigation Stack}
The ROS Navigation Stack is generic. That means, it can be used with almost any type of moving robot, but there are some hardware considerations that will help the whole system to perform better, so they must be considered. These are the requirements:
\begin{enumerate}
  \item The Navigation package will work better in differential drive and holonomic robots. Also, the mobile robot should be controlled by sending velocity commands in the form $x$, $y$ (linear velocity), $z$ (angular velocity).
  \item The robot should mount a planar laser somewhere around the robot. It is used to build the map of the environment and perform localization.
  \item Its performance will be better for square and circular shaped mobile bases.
\end{enumerate}
\begin{figure}[H]
	\centering
	\includegraphics[height=3in, width=\linewidth]{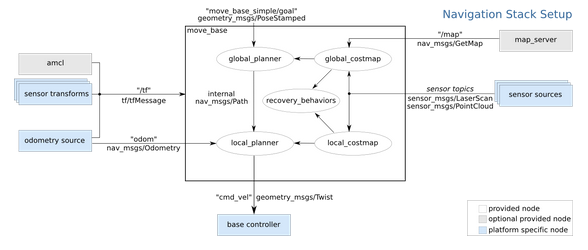}
	\caption{The navigation stack diagram\protect\footnotemark{}.}
	\label{The Navigation Stack diagram.}
\end{figure}
\footnotetext{\url{https://www.researchgate.net/figure/An-overview-of-the-ROS-Navigation-stack-8_fig1_340864490}}
According to the shown diagram, we must provide some functional blocks in order to work and communicate with the Navigation stack. Following are brief explanations of all the blocks which need to be provided as inputto the ROS Navigation stack:
\begin{itemize}
  \item Odometry source: Odometry data of a robot gives the robot position with respect to its starting position. Main odometry sources are wheel encoders, inertial measurement unit (IMU), and 2D/3D cameras (visual odometry). The odom value should publish to the Navigation stack, which has a message type of nav\_msgs/Odometry. The odom message can hold the position and the velocity of the robot.
  \item Sensor source: Sensors are used for two tasks in navigation: one for localizing the robot in the map (using for example the laser) and the other one to detect obstacles in the path of the robot (using the laser, sonars or point clouds).
  \item sensor transforms/tf: the data captured by the different robot sensors must be referenced to a common frame of reference (usually the base\_link) in order to be able to compare data coming from different sensors. The robot should publish the relationship between the main robot coordinate frame and the different sensors' frames using ROS transforms.
\item base\_controller: The main function of the base controller is to convert the output of the Navigation stack, which is a Twist (geometry\_msgs/Twist) message, into corresponding motor velocities for the robot.
\end{itemize}
\subsubsection{The move\textunderscore base node}
This is the most important node of the Navigation Stack. It's where most of the "magic" happens. The main function of the move\_base node is to move a robot from its current position to a goal position with the help of other Navigation nodes. This node links the global planner and the local planner for the path planning, connecting to the rotate recovery package if the robot is stuck in some obstacle, and connecting global costmap and local costmap for getting the map of obstacles of the environment.
The following is the list of all the packages which are linked by the move\_base node:
\begin{itemize}
  \item global-planner.
  \item local-planner.
  \item rotate-recovery.
  \item costmap-2D
\end{itemize}
The following are the other packages which are interfaced to the move\_base node:
\begin{itemize}
  \item map-server.
  \item AMCL.
  \item gmapping.
 \end{itemize}
 \begin{figure}[H]
	\centering
	\includegraphics[scale=0.55]{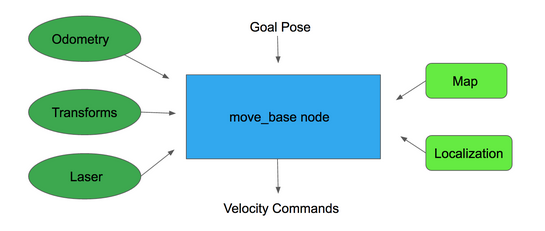}
	\caption{Move base node\protect\footnotemark{}.}
	\label{move base node.}
\end{figure}
\footnotetext{\url{https://www.theconstructsim.com/robotigniteacademy_learnros/ros-courses-library/ros-courses-ros-navigation-in-5-days/}}

\subsection{Robot Localization}
Robot localization is the process of determining where a mobile robot is located with respect to its environment. Localization is one of the most fundamental competencies required by an autonomous robot as the knowledge of the robot's own location is an essential precursor to making decisions about future actions. In a typical robot localization scenario, a map of the environment is available and the robot is equipped with sensors that observe the environment as well as monitor its own motion. The localization problem then becomes one of estimating the robot position and orientation within the map using information gathered from these sensors. Robot localization techniques need to be able to deal with noisy observations and generate not only an estimate of the robot location but also a measure of the uncertainty of the location estimate.Robot localization provides an answer to the question: Where is the robot now? A reliable solution to this question is required for performing useful tasks, as the knowledge of current location is essential for deciding what to do next.
\subsubsection{Monte Carlo Localization}
Because the robot may not always move as expected, it generates many random guesses as to where it is going to move next. These guesses are known as particles. Each particle contains a full description of a possible future pose. When the robot observes the environment it's in (via sensor readings), it discards particles that don't match with these readings, and generates more particles close to those that look more probable. This way, in the end, most of the particles will converge in the most probable pose that the robot is in. So the more you move, the more data you'll get from your sensors, hence the localization will be more precise. These particles are those arrows that are shown in RViz in the next figure.
\begin{figure}[H]
	\centering
	\includegraphics[scale=0.55]{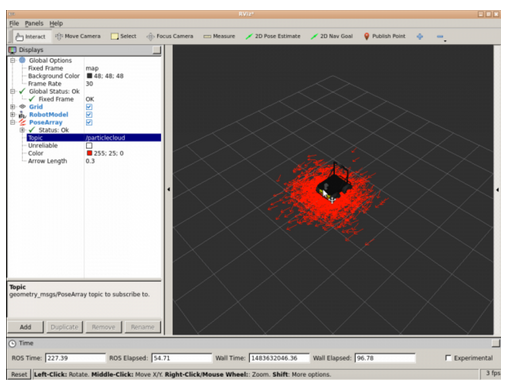}
	\caption{Monte Carlo localization.}
	\label{Monte Carlo localization.}
\end{figure}
Monte Carlo localization (MCL) \cite{fox1999monte}, also known as particle filter localization is an algorithm for robots to localize using a particle filter. Given a map of the environment, the algorithm estimates the position and orientation of a robot as it moves and senses the environment. The algorithm uses a particle filter to represent the distribution of likely states, with each particle representing a possible state, i.e., a hypothesis of where the robot is. The algorithm typically starts with a uniform random distribution of particles over the configuration space, meaning the robot has no information about where it is and assumes it is equally likely to be at any point in space. Whenever the robot moves, it shifts the particles to predict its new state after the movement. Whenever the robot senses something, the particles are resampled based on recursive Bayesian estimation, i.e., how well the actual sensed data correlate with the predicted state. Ultimately, the particles should converge towards the actual position of the robot.
\subsubsection{The AMCL Package}
In order to navigate around a map autonomously, a robot needs to be able to localize itself into the map. And this is precisely the functionality that the amcl node (of the amcl package) provides us. In order to achieve this,the amcl node uses the MCL (Monte Carlo Localization) algorithm. The AMCL (Adaptive Monte Carlo Localization) package provides the amcl node, which uses the MCL system in order to track the localization of a robot moving in a 2D space. This node subscribes to the data of the laser, the laser-based map, and the transformations of the robot, and publishes its estimated position in the map. On startup, the amcl node initializes its particle filter according to the parameters provided. Basically, the amcl node takes data from the laser and the odometry of the robot, and also from the map of the environment, and outputs an estimated pose of the robot. The more the robot moves around the environment, the more data the localization system will get, so the more precise the estimated pose it returns will be.

\subsection{Navfn}
The Navfn planner\footnote{For supplementary reading visit \url{http://wiki.ros.org/navfn}}
 is probably the most commonly used global planner for ROS Navigation. It uses Dijkstra's algorithm in order to calculate the shortest path between the initial pose and the goal pose. navfn provides a fast interpolated navigation function that can be used to create plans for a mobile base. The planner assumes a circular robot and operates on a costmap to find a minimum cost plan from a start point to an end point in a grid. The navigation function is computed with Dijkstra's algorithm, but support for an $A^{*}$ heuristic may also be added in the near future.
\subsubsection{Carrot Planner}
The carrot planner\footnote{For supplementary reading visit: \url{http://wiki.ros.org/carrot_planner?distro=noetic}} takes the goal pose and checks if this goal is in an obstacle. Then, if it is in an obstacle, it walks back along the vector between the goal and the robot until a goal point that is not in an obstacle is found. 
It, then, passes this goal point on as a plan to a local planner or controller. Therefore, this planner does not do any global path planning. It is helpful if you require your robot to move close to the given goal, even if the goal is unreachable. In complicated indoor environments, this planner is not very practical. This algorithm can be useful if, for instance, you want your robot to move as close as possible to an obstacle. But instead we use another global planner. 
\subsubsection{Global Planner}
The global planner\footnote{For supplementary reading visit: \url{http://wiki.ros.org/global_planner?distro=noetic}} is a more flexible replacement for the navfn planner. It allows you to change the algorithm used by navfn (Dijkstra's algorithm) to calculate paths for other algorithms. These options include support for
$A^{*}$ \cite{hart1972correction}, toggling quadratic approximation, and toggling grid path.
\begin{figure}[H]
	\centering
	\subfigure[Standard behavior]{
		\includegraphics[width=0.45\linewidth]{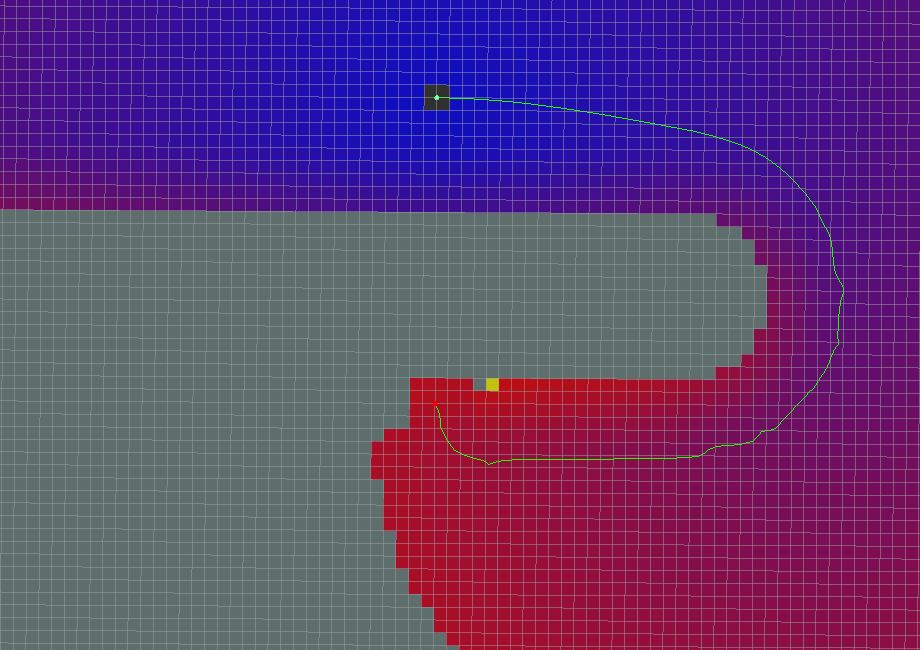}}
	\hspace{5pt}
	\subfigure[Simple potential calculation]{
		\includegraphics[width=0.45\linewidth]{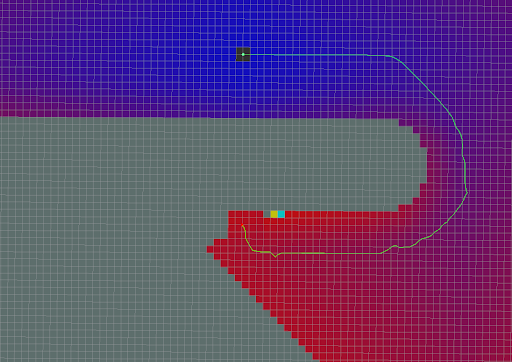}}
	\caption{Standard behavior and simple potential calculation paths\protect\footnotemark{}.}
	\label{fig:example-2}
\end{figure}
\footnotetext{\url{http://wiki.ros.org/global_planner}}

%% file: chapters/Proposed_Approach_and_Simulation_Results.tex
\chapter{Proposed Approach and Simulation Results} 
\label{ch:proposed_approach}
\section{Simulation Setup}

\subsection{ROS}
\begin{figure}[H]
	\centering
	\includegraphics[scale=0.8]{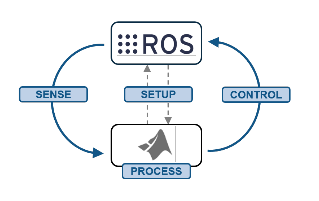}
	\caption{ROS ecosystem.}
	\label{ROS ecosystem}
	\end{figure}

ROS \cite{quigley2009ros} is a flexible platform to build robotics software applications. Its col-
lection of tools, libraries and conventions greatly simplifies the task of building
complex and robust robotics behaviours. In addition, ROS was created to encour-
age collaborative robotics software development across the world. The ecosystem of ROS is illustrated in Fig \ref{ROS ecosystem}.
The file system and nodes representation in ROS are extremely helpful in organizing and building robotics tasks.
\begin{figure}[H]
	\centering
	\includegraphics[scale=0.58]{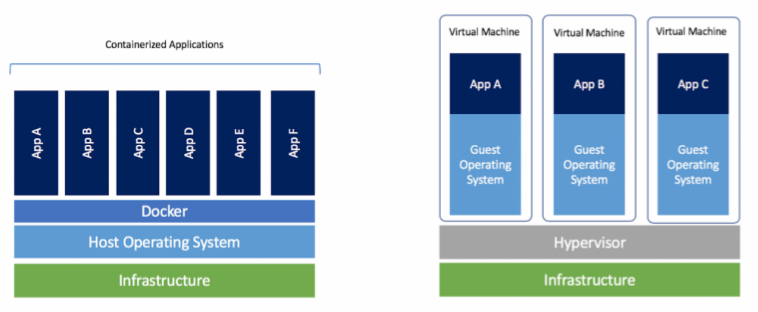}
	\caption{Difference between containers and virtual machines.}
	\label{}
	\end{figure}
ROS offers a message passing interface that
provides inter-process communication referred to as middleware. The middleware provides facilities like: publish/subscribe anonymous message passing, recording
and playback of messages, remote procedure calls, and distributed parameter
system. In addition, ROS provides common robot-specific features that help in
running basic and core robotics functions. The offered features include standard
message definitions for robots, robot geometry library, robot description language
(URDF), pose estimation and localization tools. Perhaps the most well-known
tool in ROS is Rviz. Rviz provides general purpose, three-dimensional visualiza-
tion of many sensor data types and any URDF-described robot. We can easily
visualize the laser scanned data, robot’s odometry, environment map, and many
other topics that the robot subscribes to. Rviz can be seen as a tool to visualize
what your robot can see. Another useful tool in ROS is rqt. Using the rqt graph
plugin we can introspect and visualize a live ROS system, showing nodes and the
connections between them, and being able to easily debug and understand our
running system and how it is structured.
For all the mentioned features, we have chosen ROS as a software platform
	to develop and test our robot’s navigation system on. The ROS version we use
is Kinetic Kame which is the 10th official ROS release and is supported on our
operating system, Ubuntu Xenial.
\subsection{RVIZ}
\begin{figure}[H]
	\centering
	\includegraphics[scale=0.55]{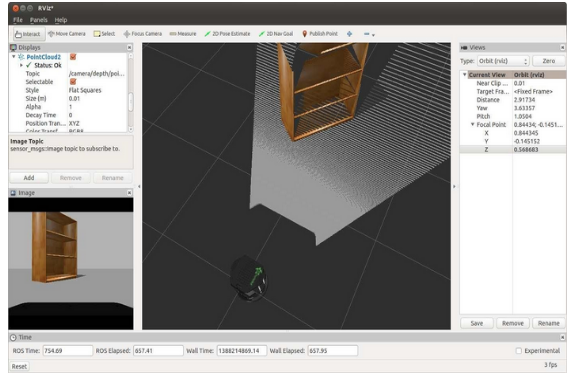}
	\caption{Sample view of \textit{rviz} configured to generate a chase perspective \cite{quigley2015programming}.}
	\label{fig:rviz}
\end{figure}

Rviz stands for ROS visualization. It is a general-purpose 3D visualization environment for robots, sensors, and algorithms. Like most ROS tools, it can be used for any robot and rapidly configured for a particular application.
rviz can plot a variety of data types streaming through a typical ROS system, with heavy emphasis on the three-dimensional nature of the data. In ROS, all forms of data are attached to a frame of reference. For example, the camera on a Turtlebot is attached to a reference frame defined relative to the center of the Turtlebot’s mobile base. The odometry reference frame, often called odom, is taken by convention to have its origin at the location where the robot was powered on, or where its odometers were most recently reset. Each of these frames can be useful for teleoperation, but it is often desirable to have a “chase” perspective, which is immediately behind the robot and looking over its``shoulders''. This is because simply viewing the robot’s camera frame can be deceiving — the field of view of a camera is often much narrower than we are used to as humans, and thus it is easy for tele-operators to bonk the robot’s shoulders when turning corners. A sample view of \textit{rviz} configured to generate a chase perspective is shown in Figure~\ref{fig:rviz}. Observing the sensor data in the same 3D view as a rendering of the robot’s geometry can make tele-operation more intuitive.

\subsection{GAZEBO}
In general, robot motions can be divided into mobility and manipulation. The mobility aspects can be handled by two- or three-dimensional simulations in which the environment around the robot is static. Simulation manipulation, however, requires a significant increase in the complexity of the simulator to handle the dynamics of not just the robot, but also the dynamic models in the scene. For example, at the moment that a simulated household robot is picking up a handheld object, contact forces must be computed between the robot, the object, and the surface the object was previously resting upon.
\begin{figure}[H]
	\centering
	\includegraphics[scale=0.55]{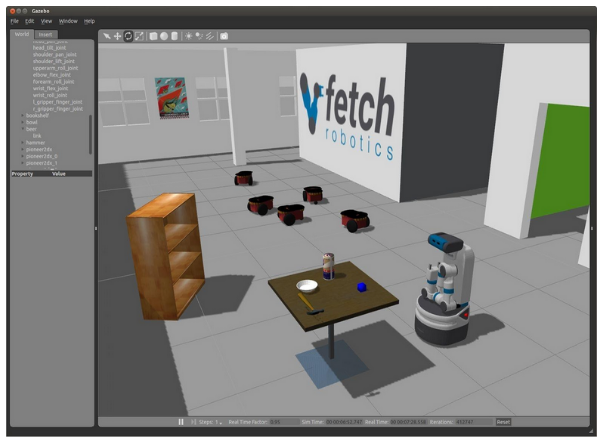}
	\caption{Gazebo \cite{quigley2015programming}.}
	\label{Gazebo.}
\end{figure}

Simulators often use rigid-body dynamics, in which all objects are assumed to be in compressible, as if the world were a giant pinball machine. This assumption drastically improves the computational performance of the simulator, but often requires clever tricks to remain stable and realistic, since many rigid-body interactions become point contacts that do not accurately model the true physical phenomena. The art and science of managing the tension between computational performance and physical realism are highly nontrivial. There are many approaches to this trade-off, with many well suited to some domains but ill suited to others.

\section{Drone Network (DroNet)}\label{DroNet}

Civilian drones are soon expected to be used in a wide variety of tasks, such as aerial surveillance, delivery, or monitoring of existing architectures. Nevertheless, their deployment in urban environments has so far been limited. Indeed, in unstructured and highly dynamic scenarios, drones face numerous challenges to navigate autonomously in a feasible and safe way. In contrast to traditional “map-localize-plan” methods.

This is achieved by DroNet: a convolutional neural network that can safely drive a drone through the streets of a city. Designed as a fast 8-layers residual network, DroNet produces two outputs for each single inputimage: a steering angle to keep the drone navigating while avoiding obstacles, and a collision probability to let the unmanned aerial vehicle (UAV) recognize dangerous situations and promptly react to them. The challenge is however to collect enough data in an unstructured outdoor environment such as a city.

\begin{figure}[H]
	\centering
	\includegraphics[scale=0.7]{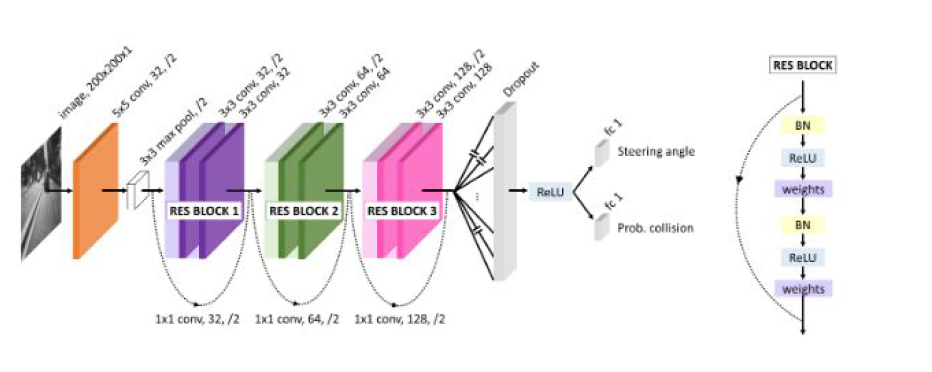}
	\caption{Dronet architecture \cite{loquercio2018dronet}.}
	\label{fig:Dronet architecture}
\end{figure}

Figure~\ref{fig:Dronet architecture}: DroNet is a forked Convolutional Neural Network that predicts, from a single $200\times200$ frame in gray-scale, a steering angle and a collision probability. The shared part of the architecture consists of a ResNet-8 with 3 residual blocks ,followed by dropout and ReLU non-linearity. Afterwards, the network branches into 2 separated fully-connected layers, one to carry out steering prediction, and the other one to infer collision probability. In the notation above, we indicate for each convolution first the kernel’s size, then the number of filters, and eventually the stride if it is different from 1.

\section{Methodology}\label{Methodology}
The approach aims at reactively predicting a steering angle and a probability of collision
from the drone on-board forward-looking camera. These are later converted into control flying commands which enable a UAV to safely navigate while avoiding obstacles. Since they aim to reduce the bare image processing time, they advocate a single convolutional neural network (CNN) with a relatively small size. The resulting network, which we call DroNet. The architecture is partially shared by the two tasks to reduce the network’s complexity and processing time, but is then separated into two branches at the very end. Steering prediction is a regression problem, while collision prediction is addressed as a binary classification problem. Due to their different nature and output range, we propose to separate the network’s last fully-connected layer. During the training procedure, we use only images recorded by manned vehicles. Steering angles are learned from images captured from a car, while probability of collision, from a bicycle.

\section{DroNet Control}
The outputs of DroNet are used to command the UAV to move on a plane with forward velocity and steering angle $\theta_k$. More specifically, they use the probability of collision $p_t$ provided by the network to modulate the forward velocity: the vehicle is commanded to go at maximal speed $v_{max}$ when the probability of collision is null, and to stop whenever it is close to 1. They use a low-pass filtered version of the modulated forward velocity $v_k$ to provide the controller with smooth, continuous inputs (0 $\leq \alpha \leq$ 1): 

   \begin{center}
      
\begin{equation} 
\label{q1}
     v_{k}=(1-\alpha)v_{k-1}+\alpha(1-p_{t})v_{max}
 \end{equation}
  
   \end{center}

Where:
\begin{itemize}
  \item $v_{k}$: The required forward velocity.
  \item $v_{k-1}$: The forward velocity from the previous iteration and zero for the first iteration.
  \item $p_{t}$: Probability of collision provided by the neural network.
  \item $v_{max}$: Max forward velocity of the robot.
\end{itemize}
similarly, they map the predicted scaled steering $s_{k}$ into a rotation around the body z-axis (yaw angle $\theta$), corresponding to the axis orthogonal to the propellers’ plane. Concretely, they convert $s_k$ from a [-1,1] range into a desired yaw angle $\theta_k$ in the range $[ -\pi /2 , \pi /2 ]$ and low pass filter it:
 
\begin{center}

\begin{equation}
\label{q2}
    \theta_{k}=(1-\beta)\theta_{k-1}+\beta(\pi/2)s_{k}
\end{equation}
    
\end{center} 

Where:
\begin{itemize}
  \item $\theta_{k}$ : The required steering angle.
  \item $\theta_{k-1}$: The steering angle from the previous iteration and zero for the first iteration.
  \item $s_{k}$: The steering angle provided by the neural network.
\end{itemize}

In all our experiments we set $\alpha = 0.7$ and $\beta = 0.5$, while $v_{max}$ was changed according to the testing environment. The above constants have been selected empirically trading off smoothness for reactiveness of the drone’s flight. As a result, they obtain a reactive navigation policy that can reliably control a drone from a single forward-looking camera. An interesting aspect of their approach is that they can produce a collision probability from a single image without any information about the platform’s speed. Indeed, they conjecture the network to make decisions on the base of the distance to observed objects in the field of view. Convolutional networks are in fact well known to be successful on the task of monocular depth estimation.

\subsection{How The DroNet Approach Implemented The Proposed Above Control in ROS}
Dronet control is implemented as anode that receives (steering angle , probability of collision ) from ``/cnn\_out/predictions'' topic which is the output of the neural network.
The neural network is presented in another node ``/dronet\_perception'' which receives images from the camera and treats it as an input to the neural network and then transmit the output (probability of collision , steering angle) to the “cnn\_ out/predictions” topic.
\subsection{Graphical Representation}
\begin{figure}[H]
	\centering
	\includegraphics[scale=0.6]{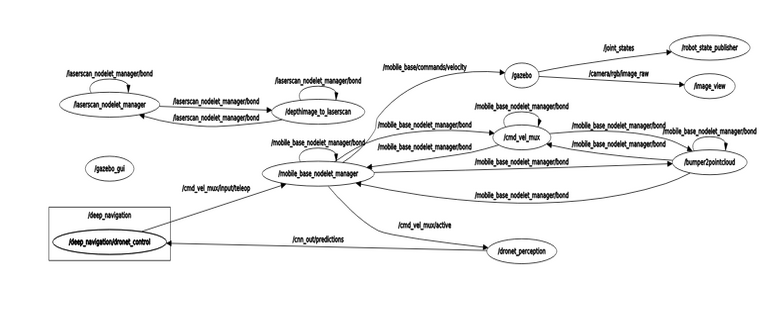}
	\caption{The rqt\_graph of dronet.}
	\label{dronet.}
\end{figure}
\section{Simulation Results for DroNet Using Ground Robot}

In this chapter, we will discuss the simulation results acquired throughout the whole project. We will divide the project into six stages and each stage will contain a result sample and description of the problem faced. The simulation tool that we are using is GAZEBO. In the following subsections, the different scenarios for DroNet ground robot based will be simulated.

\subsection {Scenario \#1}

\begin{figure}[H]
	\centering
	\includegraphics[width=1\textwidth,height=300px]{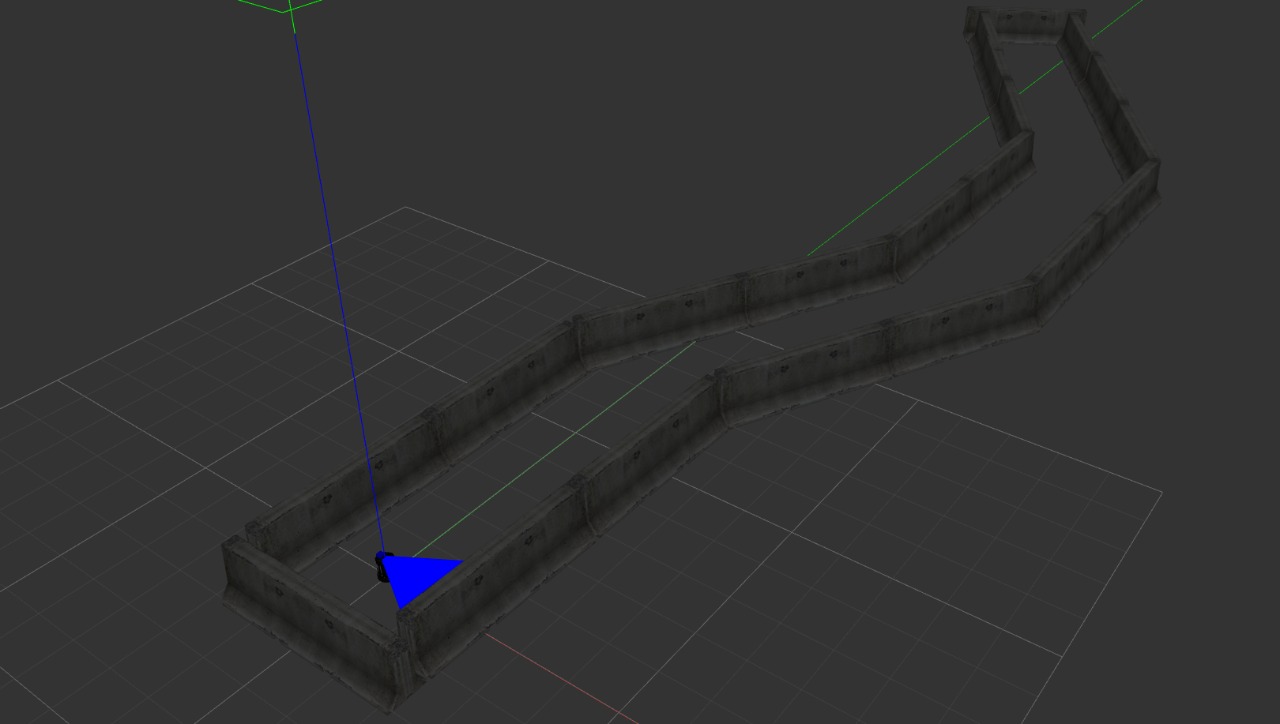}
	\caption{Single lane road environment.}
	\label{Single Lane Road.}
\end{figure}

Converting the DroNet from a drone based autonomous navigation into a ground based robot (turtlebot) based autonomous navigation. Then, creating an environment looks like the environment that the DroNet neural network was trained on. The DroNet neural network was trained on real world data targeted to navigate through streets so the simulated environment that we created is a single lane road as shown in Figure~\ref{Single Lane Road.}.

\begin{figure}[H]
	\centering
	\includegraphics[scale=0.35]{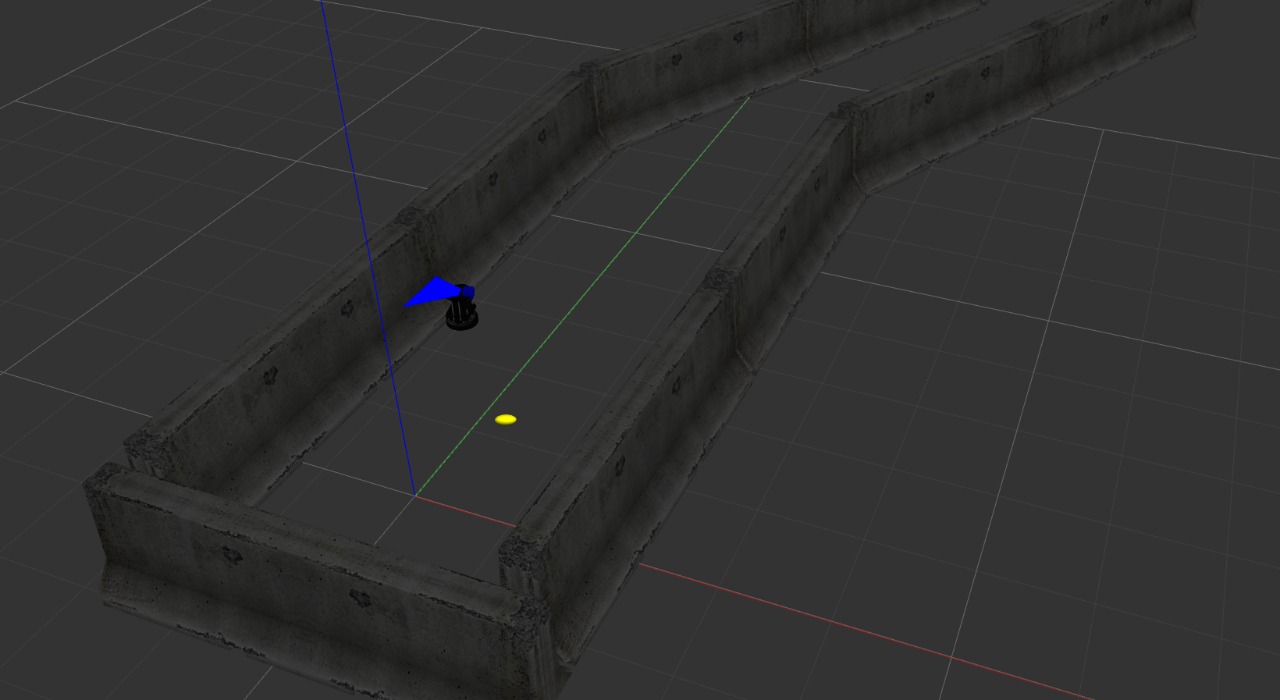}
	\caption{Robot collision.}
	\label{First Test Collision.}
\end{figure}

From Figure~\ref{First Test Collision.}, the ground robot DroNet based will collide with the wall which solved in the next subsection.

\subsection{Scenario \#2}

\textbf{\large Objective }This is achieved by tuning the LPF parameters which are linear and angular velocity in Equation \ref{q1}  and Equation \ref{q2} respectively. They set $\alpha$ = 0.3  and   $\beta$ = 0.5 while $V_{max}$ was changed according to the testing environment.

\subsection{Scenario \#3}
\begin{figure}[H]
	\centering
	\includegraphics[scale=0.45]{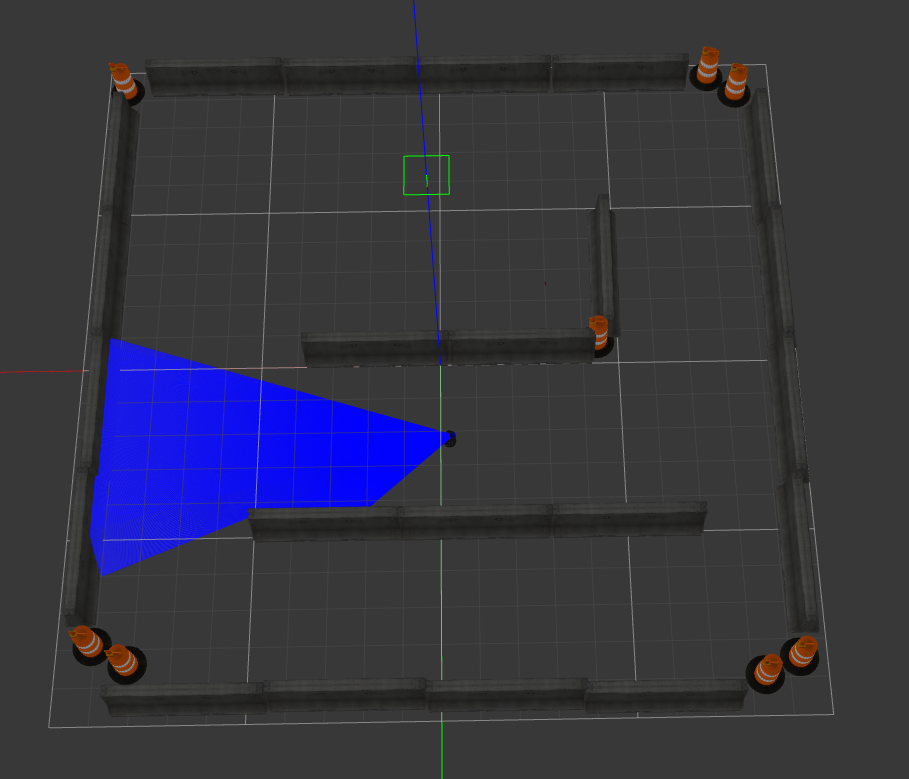}
	\caption{Indoor environment.}
	\label{3rd stage environnmet.}
\end{figure}

This scenario was concerned with trying the tuned ground based DroNet in an environment as shown in Figure \ref{3rd stage environnmet.} where the neural network wasn’t trained on it before to test how it would behave.

\begin{figure}[H]
	\centering
	\subfigure[]{
		\includegraphics[width=0.47\linewidth]{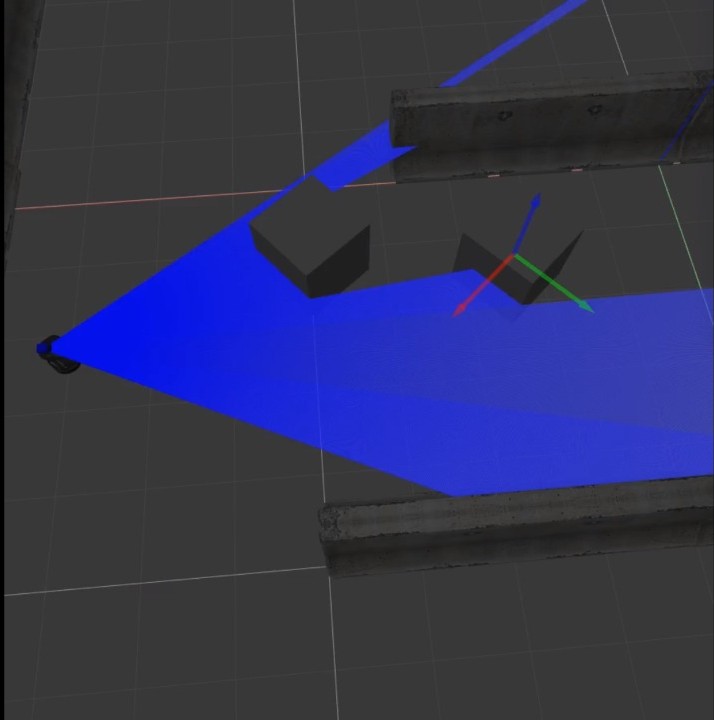}}
	\hspace{5pt}
	\subfigure[]{
		\includegraphics[width=0.47\linewidth]{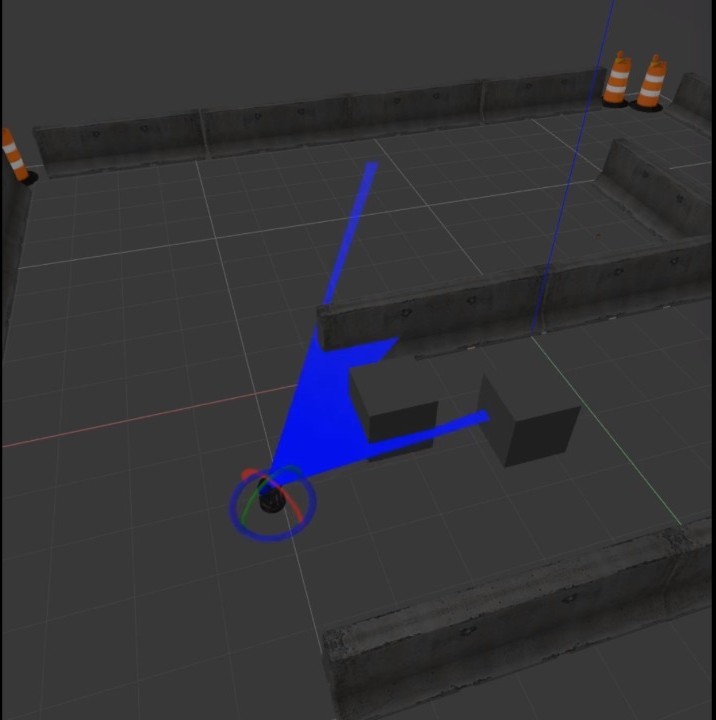}}
	\caption{(a) Start position of the robot. (b) Robot direction.}
	\label{fig:example-2}
\end{figure}

From Figure~\ref{fig:example-2} (a), we can conclude that, the blue beam from laser scanner overhead the turtleBot shows its field. The ideal scenario for the turtleBot motion is to go to the path that perfect fit the robot. But, the robot did not go through it and robot considers the whole area is blocked and turned as shown in Figure~\ref{fig:example-2} (b).

\subsection{Scenario \#4}

After retraining the neural network using a dataset generated from the same that is used in scenario \#3 as shown in Figure \ref{3rd stage environnmet.}. Dataset will be divided into inputs and outputs. Input is the images and the output is the probability of collision and steering angle corresponding to each image.
Afterwards, the collected data set is passed to the training script wich made by the DroNet's creators.

\begin{figure}[H]
	\centering
	\subfigure[]{
		\includegraphics[width=0.47\linewidth]{images/Start-Position.png}}
	\hspace{5pt}
	\subfigure[]{
		\includegraphics[width=0.47\linewidth]{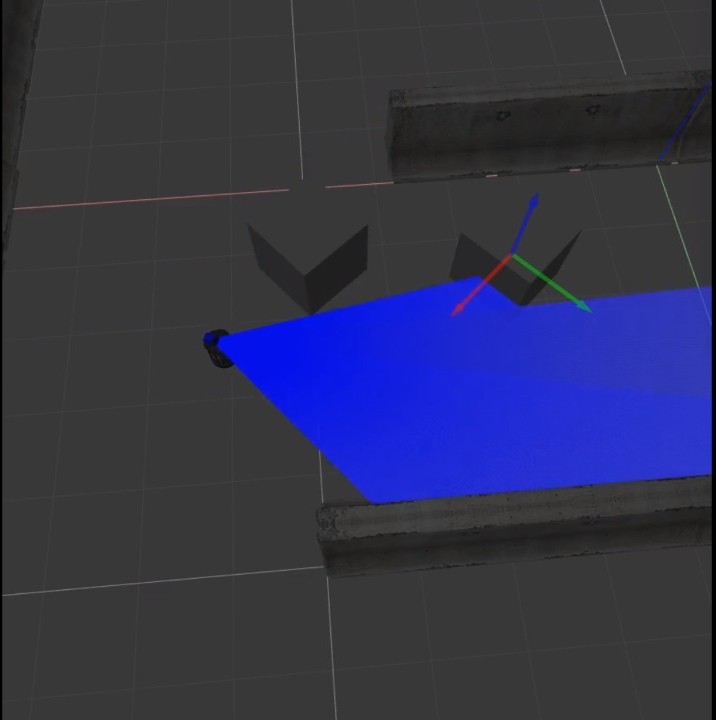}}
	\caption{(a) Start position (b) Heading to the perfect fit path.}
	\label{dd}
\end{figure}

Figure \ref{dd} (a) also shows the blue beam from laser scanner shows the  robot field. On the other hand, Figure \ref{dd} (b) shows the turtleBot moves to the path that perfectly fit the robot.


\section{How Mapping and Localization is Achieved?}

In order to perform autonomous navigation, the robot must have a map of the environment. The robot will use this map for many things such as planning trajectories, avoiding obstacles, etc. The mapping and localization is achieved as follows:

\subsection{Simultaneous Localization and Mapping}

Simultaneous Localization and Mapping (SLAM) is the name that defines the robotic problem of building a map of an unknown environment while simultaneously keeping track of the robot's location on the map that is being built.

\subsection{The Gmapping Package}

The gmapping ROS package is an implementation of a specific SLAM algorithm called gmapping (\url{https://www.openslam.org/gmapping.html}). This means that, somebody (\url{http://wiki.ros.org/slam\_gmapping})
has implemented the gmapping algorithm for us to use inside ROS, without having to code it ourself. So if we use the ROS Navigation stack, we only need to know (and have to worry about) how to configure gmapping for our specific robot (in our case Turtlebot). The gmapping package contains a ROS Node called slam\_gmapping, which allows us to create a 2D map
using the laser and pose data that your mobile robot is providing while moving around an environment. This node basically reads data from the laser and the transforms of the robot, and turns it into an occupancy grid map (OGM).

\subsection{Saving The Map}

Another of the packages available in the ROS Navigation Stack is the map\_server package. This package provides the map\_saver node, which allows us to access the map data from a ROS Service, and save it into a
file. When you request the map\_saver to save the current map, the map data is saved into two files: one is the YAML file, which contains the map metadata and the image name, and second is the image itself, which has the encoded data of the occupancy grid map.

\section{How Path Planning is Achieved?}

Moving from one place to another is a trivial task, for humans. One decides how to move in a split second. For a robot, such an elementary and basic task is a major challenge. In autonomous robotics, path planning is a central problem in robotics. The typical problem is to find a path for a robot, whether it is a vacuum cleaning robot, a robotic arm, or a magically flying object, from a starting position to a goal position safely. The problem consists in finding a path from a start position to a target position. This problem was addressed in multiple ways in the literature depending on the environment model, the type of robots, the nature of the application, etc.Safe and effective mobile robot navigation needs an efficient path planning algorithm since the quality of the generated path affects enormously the robotic application. Typically, the minimization of the traveled distance is the principal objective of the navigation process as it influences the other metrics such as the processing time and the energy consumption. Path planning is divided into global and local path planning.

\begin{figure}[H]
	\centering
	\includegraphics[scale=0.3]{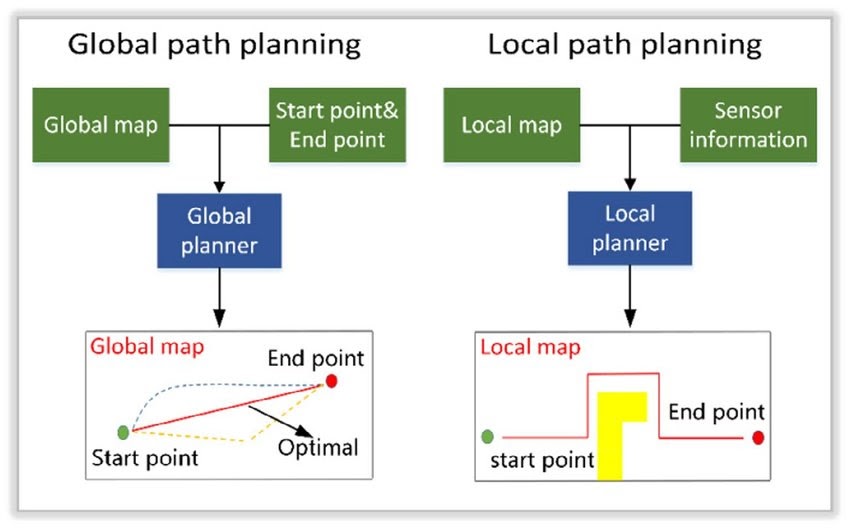}
	\caption{Global and local path planning\protect\footnotemark{}.}
	\label{global and local path planning.}
\end{figure}
\footnotetext{\url{https://www.researchgate.net/figure/Global-and-local-path-planning_fig4_322441239}}

\section{Simulation Results after Mapping and Path Planning for
Ground Robot}

In the following subsection,  the mapping and localization scenario for the turtlebot will be presented.

\subsection{Scenario \#5}

In this scenario, the turtlebot will be moving on a pre-defined path to target without needing a laser range sensor. This is achieved as the follows sequences:

\begin{enumerate}

    \item Generating an obstacle map for the environment by using the Gmapping package as shown in Figure \ref{Mapping}.
    
    \begin{figure}[H]
	\centering
	\includegraphics[width=1.0\textwidth,height=300px]{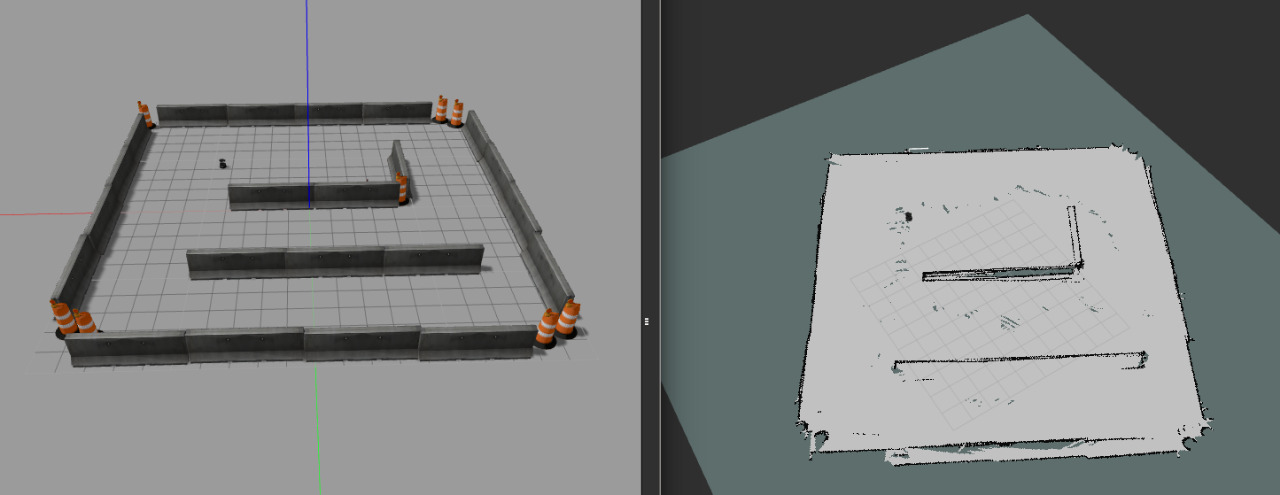}
	\caption{Mapping for the environment used.}
	\label{Mapping}
	\end{figure}
	
    \item Generating the shortest path to the target using dijkstra path planner then saving the path obtained as shown in Figure \ref{path}.
    
    \begin{figure}[H]
	\centering
	\includegraphics[width=1.0\textwidth,height=300px]{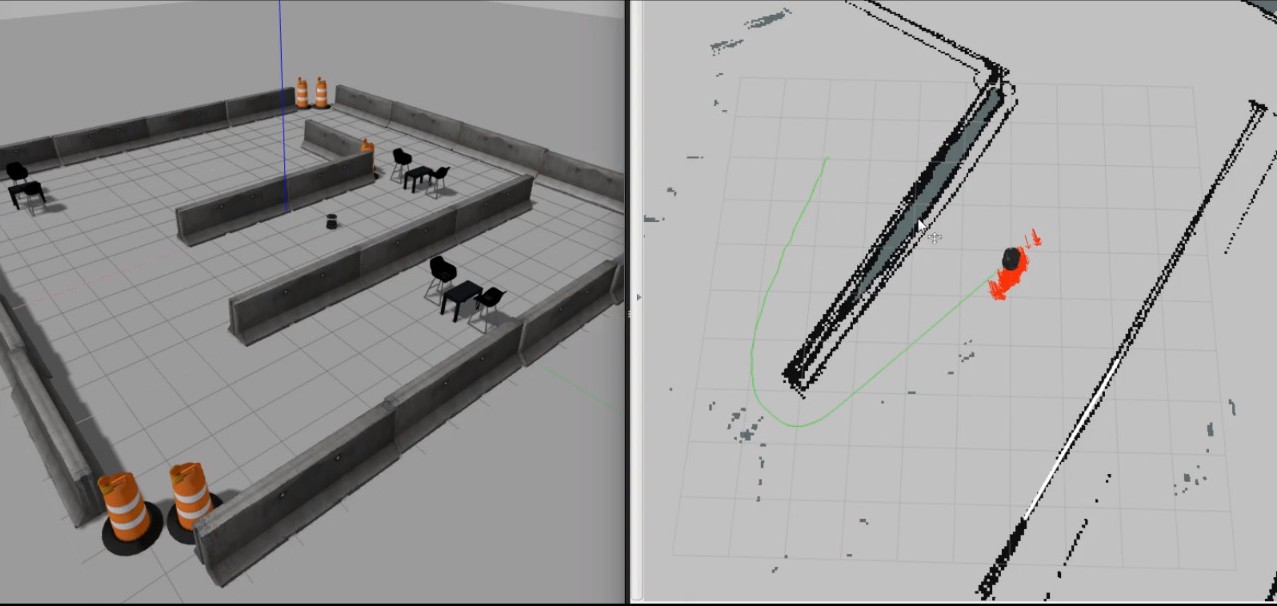}
	\caption{Path planning for the environment used.}
	\label{path}
\end{figure}

    \item Finally, creating a script that retrieve the path from the file and move on the path without Laser range sensor as shwon in Figure \ref{Without Laser Range Sensor}.
    \begin{figure}[H]
	\centering
	\includegraphics[scale=0.37]{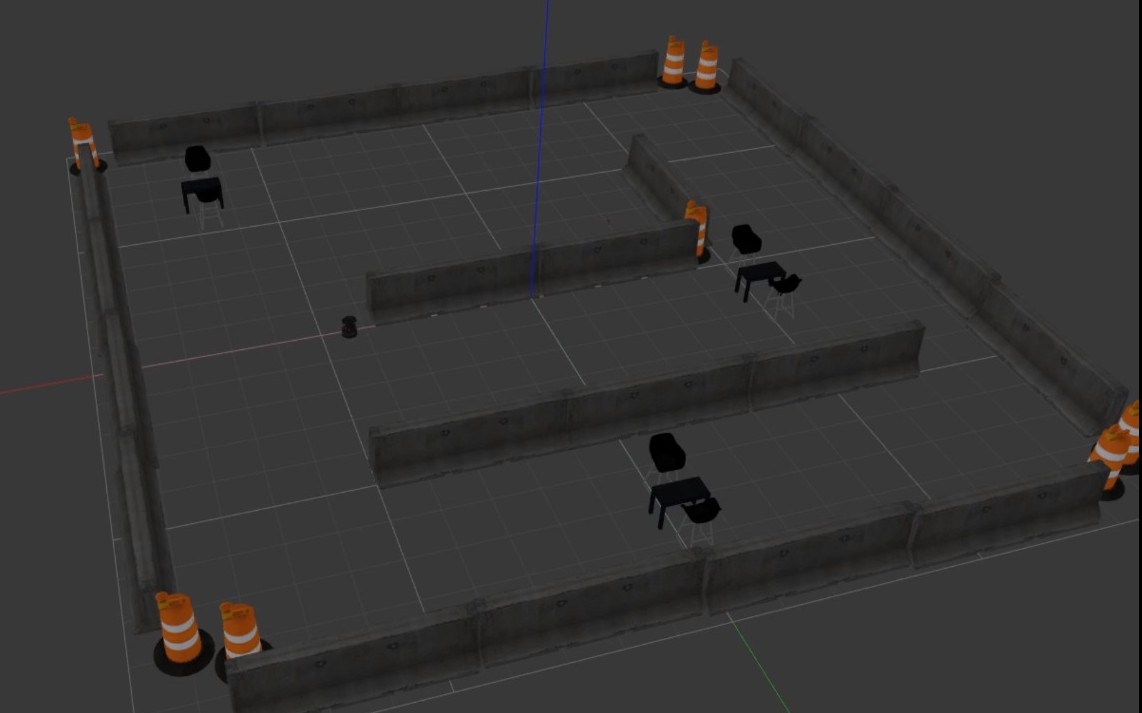}
	\caption{Moving the turtleBot on path without laser sensor.}
	\label{Without Laser Range Sensor}
\end{figure}

\end{enumerate}


\section{ Scenario \#6 Market Application}

In this scenario, the Combination of DroNet in Scenario \#4 with a script capable of retrieving a saved path and move on that path in Scenario \#5. This combination leads to a motion to a target and avoiding all dynamic object such as humans without a need of expensive laser range sensor and in replace of the laser range sensor it will use an RGP camera which is very cheap. Finally, this application can be applied in a restaurant to serve clients with low cost. 

%% file: chapters/Conclusions-and-Future-Work.tex
\chapter{Conclusions and Future Work}
\label{ch:conclusions}

\section{Conclusions}
In this project, a deep learning method that is based on a convolutional neural network is considered. Software simulation tools have been learned and used to achieve the main objectives of our project such as Linux, robot operating system (ROS), C++, python, and GAZEBO simulator. 
Software simulation is to achieve the output of the method using the laser sensor data that is preprocessed in such a way that enables the network to decide which direction to follow to move nearer to the target. 
The goal-oriented motion problem for the DroNet approach has been solved using mapping and path planning. In addition to this, this thesis proposed a marketing service restaurant application at a low cost. Finally, the simulation results are very promising and the robot performance is good.

\section{Future Work}

Now, we are going to list some possible directions for future work:
\begin{itemize}
  \item We first realize our project by hardware implementation.
  \item This work may be extended for agriculture application by combination internet of thing (IOT) approach.
  \item The goal oriented motion problem for the DroNet approach my solved via potential field approach.
  
\end{itemize}